\documentclass[10pt,twocolumn,letterpaper]{article}

\usepackage[pagenumbers]{iccv}
\usepackage{enumitem}
\usepackage{paralist}
\usepackage{cuted}
\usepackage{multirow}
\usepackage{lipsum}

\definecolor{iccvblue}{rgb}{0.21,0.49,0.74}
\usepackage[pagebackref,breaklinks,colorlinks,allcolors=iccvblue]{hyperref}
\usepackage{duckuments}
\usepackage{boldline}
\usepackage{booktabs}
\usepackage{dcolumn}
\usepackage{multirow, makecell, caption}
\usepackage{colortbl}
\usepackage{amssymb}
\usepackage{pifont}
\usepackage{array}
\usepackage{multirow}
\usepackage{array}
\usepackage{graphicx}  
\usepackage{arydshln}
\usepackage{algorithm}
\usepackage{listings}

\definecolor{iccvblue}{rgb}{0.21,0.49,0.74}

\def\paperID{3189} 
\def\confName{ICCV}
\def\confYear{2025}

\title{Open-ended Hierarchical Streaming Video Understanding 

with Vision Language Models}

\author{
Hyolim Kang$^*$ \quad Yunsu Park$^*$ \quad Youngbeom Yoo \quad  Yeeun Choi \quad Seon Joo Kim   \\
Yonsei University\\
\texttt{\{hyolimkang,ysp7954,youngbeom.yoo,yeeun-choi,seonjookim\}@yonsei.ac.kr}
}

\begin{document}
\twocolumn[{%
\maketitle
\vspace{-11mm}
\begin{center}
\footnotesize{\url{https://sites.google.com/view/yunsupark/projects/openhouse}}
\end{center}
}]
\renewcommand{\thefootnote}{\fnsymbol{footnote}}
\setcounter{footnote}{1}
\footnotetext[1]{Equal contribution.}
\renewcommand{\thefootnote}{\arabic{footnote}}
\begin{abstract}
We introduce Hierarchical Streaming Video Understanding, a task that combines online temporal action localization with free-form description generation. 
Given the scarcity of datasets with hierarchical and fine-grained temporal annotations, we demonstrate that LLMs can effectively group atomic actions into higher-level events, enriching existing datasets.
We then propose OpenHOUSE (Open-ended Hierarchical Online Understanding System for Events), which extends streaming action perception beyond action classification. 
OpenHOUSE features a specialized streaming module that accurately detects boundaries between closely adjacent actions, nearly doubling the performance of direct extensions of existing methods.
We envision the future of streaming action perception in the integration of powerful generative models, with OpenHOUSE representing a key step in that direction.

\end{abstract}    
\vspace{-3mm}
\section{Introduction}
\label{sec:intro}

Imagine a cyclist wearing an egocentric camera while repairing a bicycle.
If we are designing a system to assist the cyclist, it must not only recognize the fine-grained action of ``tightening a bolt,'' but also understand that it is a substep of ``changing the wheel,'' which in turn contributes to the larger goal of ``repairing the bike.'' 
Recognizing the importance of the hierarchical structure, we argue that such understanding must be acquired instantaneously for effective real-time assistance.
Given this complexity, relying on predefined class categories is suboptimal, as their limited expressiveness restricts the system’s ability to capture the hierarchical structure in detail.

\begin{figure}
    \centering
    \includegraphics[width=\columnwidth]{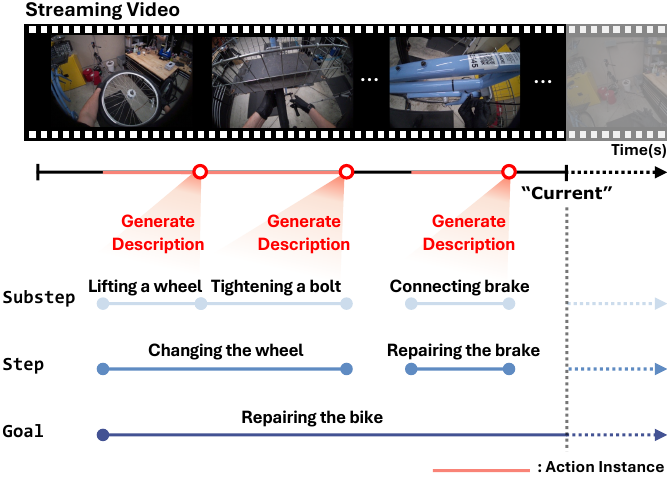}
    \caption{Illustration of the Hierarchical Streaming Video Understanding task. 
    When action instances are detected, the corresponding hierarchical descriptions are generated instantly. The hierarchy consists of three levels: $Goal$, $Step$, and $Substep$. The $Goal$ description can be generated before the entire video concludes.}
    \label{fig:teaser}
    \vspace{-5mm}
\end{figure}

This highlights the need for an instantaneous, open-ended and hierarchical understanding of streaming input (Figure \ref{fig:teaser}).
Large-scale Vision Language Models (VLMs) \cite{clip, vificlip, gpt4}, with their powerful reasoning capabilities, are well-suited for handling this challenge. 
However, applying VLMs to continuous visual streams presents unique challenges beyond standard offline video understanding \cite{timechat, videochatgpt}.
First, the model must determine ``when'' to generate output within the task hierarchy. 
Continuously calling the VLM is computationally expensive, and generating descriptions for every frame would result in excessive, fragmented, and inaccurate descriptions.
Second, given the untrimmed, extensive nature of the video, the model must determine which segment is relevant to the ongoing action, as processing the entire history is impractical.
Finally, using this information, the model should provide an immediate description of the ongoing action.

Rather than devising a single model, we decouple the task into two parts: (i) a lightweight ``streaming'' perception module that processes each frame and performs hierarchy-aware action detection in real time, calling the VLM only when needed, and (ii) a VLM that generates descriptions only when called by the perception module.
This disentanglement offers distinct advantages over a single model approach. 
First, it enables efficient operation by continuously running the lightweight module and sparingly activating the computationally expensive VLM. 
Second, separating streaming perception tasks from caption generation allows the VLM to function as originally trained, enabling zero-shot inference.
By avoiding fine-tuning or task-specific adaptation, we preserve the VLM's generalization ability and avoid the degradation typically caused by tuning on relatively small custom datasets.

The proposed system, \textbf{Open}-ended \textbf{H}ierarchical \textbf{O}nline \textbf{U}nderstanding \textbf{S}ystem for \textbf{E}vents (\textbf{OpenHOUSE}), is an approach that bridges the gap between established online action detection methods and large-scale VLMs.
Without compromising the strict online nature of existing online action detection formulation, we effectively incorporate the generative capabilities of large-scale VLMs by employing a lightweight online model as a \textit{VLM caller} and an \textit{online hierarchical action segmentor} to determine both which parts of the streaming video should be processed and their hierarchical structure.
This design distinguishes our approach from previous methods that adopted VLMs for streaming perception, such as relying on fixed decoding points \cite{streamingvideocaptioning} or invoking the VLM for every incoming frame \cite{VideoLLM-online}.

However, a straightforward integration of existing online modules with large-scale VLMs proves insufficient due to the following challenges:

\noindent \textbf{Scarcity of hierarchical annotations:} While VLMs do not require fine-tuning to grasp hierarchical structures, detailed temporal annotations that include hierarchical relationships are essential for training a streaming module to understand action hierarchies.
However, although many datasets \cite{epickitchen, egoexo4d, egoProceL, assembly} offer atomic action annotations, hierarchical annotations remain scarce.
To address this, we propose a dataset generation pipeline that leverages the reasoning capacity of LLMs to group atomic actions into higher-level hierarchical structures.
Notably, experiments on Ego4D-GoalStep \cite{ego4dgoalstep}, which has ground-truth hierarchical annotations, show that models trained on the generated dataset perform competitively with those trained on ground truth, confirming the reliability of our dataset generation pipeline. 
By training the streaming module with these annotations, our dataset pipeline effectively distills LLM's hierarchical awareness to the streaming module.

\noindent \textbf{Class-agnostic action boundary modeling:}
Similar to class-agnostic object detectors in open-vocabulary recognition \cite{rfcn, fast-rcnn, efficientdet}, the streaming module in OpenHOUSE operates without predefined class labels. 
However, previous class-agnostic action boundary detection strategy—based on detecting transitions from background to action frames—significantly struggles in procedural videos due to the lack of sufficient background frames between actions.
To address this, we propose a new strategy that combines estimated action presence (or ``actionness'') with progression to identify action start and end points.
This approach has proven highly effective, nearly doubling the F1 score compared to conventional actionness-based methods \cite{miniroad, testra} in the OpenHOUSE framework.

\noindent Our contributions can be summarized as follows:
\begin{compactitem}
    \item We introduce the new, practical problem of Open-ended Hierarchical Streaming Video Understanding and propose \textbf{OpenHOUSE} as a solution.
    \item We validate the capability of LLMs to group atomic actions into higher-level annotations, addressing the scarcity of hierarchical temporal data.
    \item We propose a novel strategy for detecting action boundaries, capable of handling instances that are strictly adjacent without background frames.
    \item We conduct extensive experiments including the cross-dataset evaluation, which demonstrates the generalizability and scalability of our approach.
\end{compactitem}


\section{Related Work}

\subsection{Streaming Action Perception}
Online Action Detection (OAD) \cite{oad} is a well-established task in streaming video understanding that involves identifying the action class of the current frame in a streaming setting. While extensive research has focused on OAD \cite{trn, idn, colar, oadtr, lstr, gatehub, testra, miniroad, openvocab-oad}, its frame-centric approach falls short for real-world streaming video analysis, which requires event-level recognition.

Conversely, Online Temporal Action Localization (On-TAL) \cite{simon, actionswitch, online-hat, online-mat, 2pesnet, oat}, the task recently introduced in \cite{cagqil}, targets real-time event detection. 
Beyond the constraint of unobserved future frames, On-TAL requires immediate responsiveness, producing detection results as soon as an event terminates.
This unique requirement makes the problem more challenging, preventing the naive extension of previous approaches to On-TAL.

\noindent\textbf{Our Work}
Targeting practical online scenarios, we adopt the OAD-based On-TAL setting \cite{cagqil}.
While a relevant recent study \cite{openvocab-oad} explores VLMs for per-frame action classification, we focus on leveraging VLMs' free-form generation and reasoning capabilities to capture the hierarchical structure of ongoing actions.

\subsection{Streaming Open-ended Video Understanding}

\noindent \textbf{Video Captioning}
Many works have focused on generating captions that describe video content. Notably, SDVC \cite{streamingvideocaptioning} adopted the concept of a “decoding point,” which enables captions to be generated without observing the full video.
However, unlike OAD or On-TAL, SDVC is not designed for prompt responses in online scenarios. Its primary innovation lies in enabling caption generation without full video access, rather than recognizing ongoing actions and responding in real time.
\begin{figure}
    \centering
    \vspace{-8mm}
    \captionsetup{aboveskip=0pt}  %
    \captionsetup{belowskip=0pt}  
    \includegraphics[width=\columnwidth]{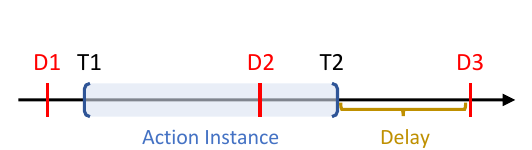}
    \caption{In the SDVC setting, action instances are generated at fixed decoding points (e.g., D3), which occur at regular intervals, inevitably introducing a delay (D3 - T2). In contrast, the On-TAL formulation requires action instance detection at T2, ensuring real-time responsiveness. A systematic analysis of this delay is provided in Sec. \ref{sec:results_and_analysis}.}
    \label{fig:sdvc-comp}
    \vspace{-6mm}
\end{figure}

\noindent \textbf{Video Question Answering (VQA)}
The application of VLMs for VQA \cite{videochatgpt} has gained increasing attention due to their strong reasoning abilities. While many studies address conventional, “offline” VQA, the streaming VQA scenario—despite its practical importance—has been relatively underexplored. Recent works \cite{VideoLLM-online, videollm-online-mod} have enabled streaming VQA by fine-tuning VLMs with timestamp-aligned custom VQA datasets. However, these frameworks require the fine-tuned VLM to perform inference on \textit{every incoming frame} and to handle all temporal recognition tasks and text generation by one model.

\noindent \textbf{Our Work}
Following the On-TAL problem definition \cite{cagqil}, our approach runs in a strict online setting: open-ended descriptions must be generated instantly in response to the current action status, whereas in \cite{streamingvideocaptioning}, decoding points are not required to align with ongoing actions (Figure \ref{fig:sdvc-comp}).
Furthermore, by decoupling temporal understanding from text generation, OpenHOUSE achieves a more sophisticated, hierarchical temporal comprehension, enabling it to capture structured \textit{action intervals} rather than aligning predictions to single points in time \cite{VideoLLM-online, videollm-online-mod}.

\subsection{Hierarchical Video Understanding}
Hierarchical video understanding seeks to recognize actions across multiple levels of granularity, from fine-grained actions to higher-level activities. Although this hierarchical structure aligns with natural action dynamics, few recent works address this topic \cite{videorecap, ego4dgoalstep}.

\noindent\textbf{Our Work}
To the best of our knowledge, OpenHOUSE is the first framework to explore hierarchical understanding in a streaming setting. Unlike \cite{videorecap}, which focuses on generating high-quality captions for a given clip or segment by recurrently leveraging low-level captions, our approach emphasizes identifying the ``temporal boundaries'' of action instances across different hierarchical levels.

\section{Target Task and Our Approach}

\subsection{Hierarchical Streaming Video Understanding}
\label{sec:problem_def}
Consider an untrimmed video $V\!=\!\{x_{\tau}\}^{T}_{\!\tau=1}$ with $M$
hierarchical action instances $\Psi\!=\!\{\psi_m\}_{m=1}^M\!=\!\{(t^s_m, t^e_m, d_m, h_m)\}_{m=1}^M$ provided in a streaming format. 
Here, $x_{\tau}$ denotes the $\tau$th frame; $t^s_m\in \mathbb{R}$, $t^e_m\in \mathbb{R}$ are the start and the end timestep; and $d_m$ and $h_m \in \mathbb{N}$ represent the (possibly free-form) description and hierarchical level of the $m$th action instance $\psi_m$ respectively.
To ensure immediate action recognition, the system must respond promptly to action termination. 
We therefore adopt the OAD based On-TAL formulation \cite{cagqil, simon, actionswitch} for this task.
Note that recent TAL-based On-TAL formulations \cite{oat, online-hat, online-mat} relax online constraints through anchor-based boundary regression, allowing some flexibility in the timing of action instance generation, which can result in either early or delayed predictions.
In contrast, the OAD-based formulation adheres to a per-frame grouping paradigm that strictly aligns action termination with the current frame, which is why we opted for this setting.

Our objective is to generate and accumulate action proposals $\psi$ as soon as each action completes, reconstructing $\Psi$ \textit{without retrospective modification} to any $\psi$.
This work focuses on a three-level hierarchy: $substep, step, goal$, corresponding to $h= 1, 2, 3$ respectively.
$substep$ and $step$ are defined at the instance level with temporal annotations, while the $goal$ applies to the entire video.

\begin{figure}
    \centering
    \includegraphics[width=\columnwidth]{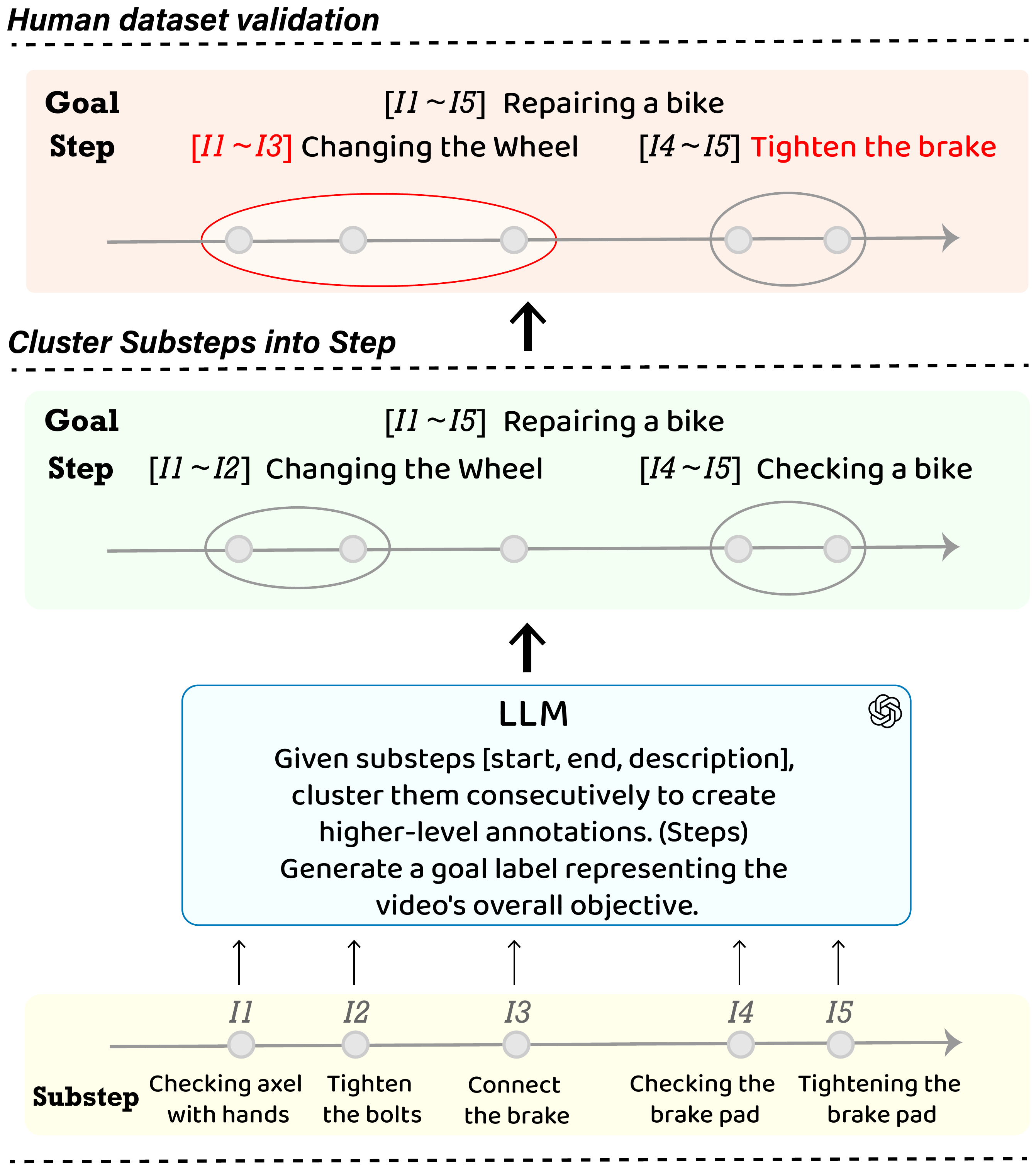}
    \caption{Dataset generation pipeline. Given substep instances from $I1$ to $I5$, the LLM clusters these substeps into three steps, generating step descriptions, timestamps, and a goal description. Afterward, human validation is conducted to include the missing annotation $I3$ into the step annotations of $I1$ to $I2$, and to revise the step description for $I4$ to $I5$.}
    \label{fig:dataset-pipeline}
    \vspace{-6mm}
\end{figure}

\begin{figure*}
    \centering
    \includegraphics[width=\linewidth]{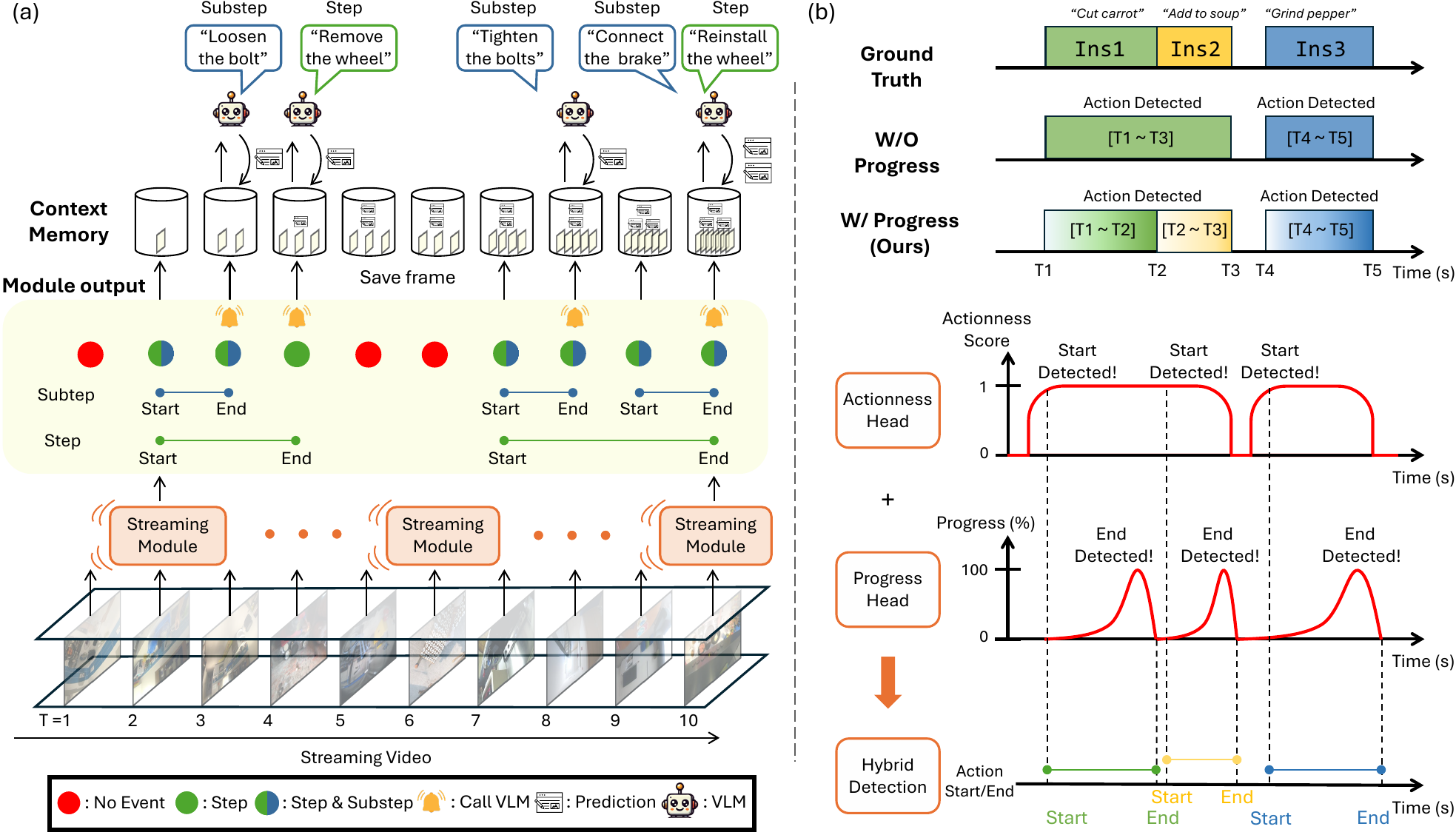}
    \caption{(a) Illustration of OpenHOUSE.
    The lightweight Streaming module processes every frame, while VLM inference is triggered selectively at $T=3,4,8,10$, indicated by the yellow bell icon.
    (b) Diagram depicting how our hybrid action boundary detection works.
    For instance, at $T2$, a sudden drop in the progression signals the action end, marking the current frame as a background frame. 
    In the next timestep, a high actionness score ($=1.0$) classifies the frame as an action frame, resulting in a background-to-action transition and marking it as the start of \texttt{Ins2}.}   
    \label{fig:method}
    \vspace{-3mm}
\end{figure*}

\subsection{Dataset Pipeline}
To enable the streaming module to recognize hierarchical structures, both hierarchical action annotations and fine-grained temporal annotations are essential.
While some datasets offer either fine-grained atomic action annotations \cite{epickitchen, egoexo4d} or hierarchical captions \cite{videorecap}, very few provide both. 
To the best of our knowledge, the Ego-4D Goalstep \cite{ego4dgoalstep} dataset is the only hierarchical action recognition dataset with sufficiently fine-grained temporal annotations.
To address this limitation, we leverage the reasoning capacity of LLMs.
Just as humans can intuitively group atomic action annotations into higher-level categories, we utilize an LLM to perform this hierarchical grouping automatically.

Figure \ref{fig:dataset-pipeline} illustrates the overall framework of our dataset generation pipeline, which consists of two main stages: clustering substeps and human validation. 
In the clustering substeps stage, we employ an LLM to group atomic actions, or $substeps$, into meaningful clusters, referred to as $steps$. 
After this initial clustering, a post-processing step is applied to further refine the results.
Subsequently, the LLM generates appropriate summaries for each set of substeps to form the step-level annotations, and also produces goal-level annotations that best capture the overall purpose of the video.
For more organized descriptions, we applied K-means clustering to group similar steps (200 for Ego-Exo4D Keystep~\cite{egoexo4d} and 500 for Epic-Kitchens 100~\cite{epickitchen}). 
The LLM then generated representative captions for each group to categorize the step annotations.

In the human validation stage, generated annotations are evaluated and adjusted across two aspects: temporal consistency and caption quality.
Temporal consistency involves checking missing instances, as well as addressing abnormally long or short temporal annotations.
For caption quality, human evaluators revise step-level captions that fail to adequately reflect the content of their substeps, fill in missing captions, and adjust goal captions when they do not accurately convey the video's overall purpose.
Empirically, we found that \textit{minimal} adjustment is needed in this stage due to the powerful reasoning capabilities of LLMs.

Using this pipeline, we successfully augmented the Ego-Exo4D Keystep~\cite{egoexo4d} dataset and the Epic-Kitchens 100~\cite{epickitchen} dataset into hierarchical formats, thereby addressing the scarcity of hierarchical annotations.
In Section \ref{sec:experiment}, we further demonstrate the validity of our generated annotations.
These annotations will be made public along with the code.

\subsection{Approach: OpenHOUSE}
\noindent\textbf{Overview}
Figure \ref{fig:method} (a) illustrates the schematic diagram of \textbf{Open}-ended \textbf{H}ierarchical \textbf{O}nline \textbf{U}nderstanding \textbf{S}ystem for \textbf{E}vents (\textbf{OpenHOUSE}), which consists of three main components: the Streaming module, Context memory, and the frozen VLM. The Streaming module processes each incoming frame, performing class-agnostic On-TAL based on OAD \cite{actionswitch}. As an extended OAD model, this module also tracks ongoing action instances at different hierarchical levels, determining whether to store the incoming frame in the context memory.

When the module detects the termination of an action instance, it returns a class-agnostic action instance along with its hierarchy level. 
This action instance serves as a ``query'' to retrieve relevant information from the context memory, such as frames and previous predictions.
The retrieved information, consisting of relevant frames and previous text-based predictions, is then passed to the frozen VLM, along with prompts tailored to the action instance's hierarchy level, to generate an open-ended description of the segment. 
The generated prediction is subsequently added to the context memory for future reference.
The key feature is that the lightweight Streaming module processes every frame, while the computationally heavy VLM is used sparingly—resulting in an efficient setup for online scenarios.

\subsubsection{Streaming Module}
\noindent\textbf{Motivation} The Streaming module in OpenHOUSE performs class-agnostic On-TAL, but faces unique challenges in identifying step and substep boundaries.
Prior approaches for class-agnostic On-TAL \cite{cagqil, actionswitch} detect action instances by grouping action frames—a straightforward method when action instances are sparsely distributed with sufficient background frames naturally separating them
(\texttt{Ins2} and \texttt{Ins3} in Figure \ref{fig:method} (b)).
By simply detecting transitions between background frame and action frames, these methods can identify action boundaries effectively.

Nevertheless, in procedural videos, which are the primary focus of our work, action instances often occur in close succession with minimal or no background frames in between (\texttt{Ins1} and \texttt{Ins2} in Figure \ref{fig:method} (b)).
Following prior methods in this context leads to a monotonous sequence of action frames, interpreting multiple instances as a single continuous instance 
(\texttt{w/o Progress} in Figure \ref{fig:method} (b)).
Even when few background frames exist between action instances, distinguishing them is still challenging.
Ideally, both action ends and starts should be marked as action frames, with only the short background marked as background—a distinction that is theoretically possible but extremely difficult in practice.
This issue is unique to the class-agnostic setting; in class-aware settings, boundaries can be clearly marked by changes in action labels \cite{progress-aware}, but such cues are absent in the class-agnostic setting.

\noindent\textbf{Hybrid action boundary detection} 
To address this issue, we introduce a novel strategy for determining action boundaries.
In addition to measuring actionness, we also measure the ``progress'' of action instances and detect sudden drops in this progress to signify action ends. 
Note that ground-truth progress can be derived from temporal annotations (e.g., if an action occurs between 2s and 4s, the timestamp at 3s corresponds to 50\% progress), and it is used for training progression heads.
Unlike prior methods that use progress as auxiliary information \cite{progress-aware}, we explicitly leverage the estimated progress score for action boundary detection.

However, using progress alone for action boundary detection is not feasible, as it increases gradually at the start, providing no distinctive signal for detecting action start. 
Thus, we adopt a hybrid approach: action starts are detected using conventional actionness-based methods, while action ends are identified by monitoring progress drops.

Figure \ref{fig:method} (b) illustrates our action boundary detection algorithm within a single hierarchy. 
Action starts and ends are detected separately by the actionness head and progress head, respectively.
For multiple hierarchies, this approach can be extended by assigning additional progress head to each hierarchy and incorporating a state-emitting head \cite{actionswitch} instead of binary actionness head.

\noindent\textbf{Implementation Details}
Following recent OAD literature \cite{miniroad}, we use a simple RNN backbone shared across three heads: a state-emitting head, a progression head for steps, and a progression head for substeps.
The state-emitting head is trained using standard cross-entropy loss, as in \cite{actionswitch}.
Although the progression heads are essentially regression tasks, we apply histogram loss \cite{stopregressing, histogramloss} to transform them into classification problems, allowing all heads to be trained with cross-entropy loss.

\subsubsection{VLM}
\label{sec:vlm}
Exploiting the fact that our Streaming module sparingly generates action instances, we use each instance generation as a signal to call the VLM, enabling occasional VLM inference. 
In this setup, the VLM is solely responsible for generating captions based on provided frames and text, with no special online behavior required. 
This allows us to leverage the VLM's strong zero-shot inference capability, which enables the VLM to remain frozen throughout training.
Additionally, inspired by RAG \cite{rag}, we incorporate information from previous predictions instead of relying solely on the current action instance, assuming a context memory that stores relevant past frames and predictions.

For example, assume that the Streaming module detects the end of a $step$. 
Then, the module will provide the action start and current timestamps along with the corresponding hierarchy level: $step$.
Using these, relevant data—such as previous $step$ predictions and frames of detected $substeps$ within that interval—is retrieved from the context memory. Substep frames are sampled from these relevant $substeps$ instead of using uniform sampling, which could overlook important details. 
This, along with a customized prompt for the $step$ hierarchy, is passed to the VLM, producing appropriate description. The prediction is then added to the context memory for future use.
Exact prompts and other details are provided in the supplementary materials.

\section{Experiment and Analysis}
\label{sec:experiment}
\subsection{Experimental Setup}
\noindent\textbf{Datasets} 
We begin our evaluation with Ego4D-GoalStep (EgoGS) \cite{ego4dgoalstep}, which includes ground truth hierarchical temporal annotations. 
After validating our dataset generation pipeline, we extend our experiments to Ego-Exo4D Keystep (EgEx) \cite{egoexo4d} and Epic-Kitchens 100 (EK100) \cite{epickitchen}. 
Detailed dataset analyses, including statistics, are provided in the supplementary materials.
Unless specified, hereafter ``EgEx'' refers to Ego-view in Ego-Exo4D Keystep dataset.

\noindent\textbf{Features}
Following established conventions in video temporal understanding \cite{actionformer, bmn, oad, online-hat, online-mat}, we use a pretrained vision encoder to extract features from raw frames.
Except for the exo view of EgoExo4D, which uses the official features, we use the Egovideo \cite{egovideo} encoder. 
All experiments, except for the exo view of EgoExo4D, are carried out with the \textit{same features} to ensure fair comparison.
Note that features are extracted online, and we take this into account when measuring inference speed.

\noindent\textbf{Evaluation Metric}
Our evaluation protocol focuses on three aspects: (i) class-agnostic localization ability of the Streaming module, (ii) quality and alignment of generated descriptions, and (iii) joint assessment of both criteria.

Based on the problem formulation (Section \ref{sec:problem_def}), our method is an open-ended extension of OAD based On-TAL.
Instead of simply predicting class indices, it evolves to generate open-ended descriptions of the target action instance.
Thus, following a recent work \cite{actionswitch, vid2seq, streamingvideocaptioning}, we opt for F1@$tIOU$ as our main metric to evaluate the class-agnostic localization performance of the Streaming module.

For description quality and alignment, we use GPT-Score, a standard GPT-assisted evaluation for open-ended VLM outputs~\cite{moviechat, chatunivi, pllava, stllm, etbench}. Following \cite{videochatgpt}, we score all true positive predictions on a 1–5 scale across multiple criteria and report the average GPT-Score in experiments.
As noted in \cite{minigpt}, we also found that conventional N-gram metrics (e.g., CIDEr \cite{cider}) are not the optimal choice for zero-shot VLM evaluation.

To jointly evaluate localization ability and description alignment, we adopt a zero-shot evaluation strategy from CLIP \cite{clip}, encoding the predicted and all ground truth descriptions into vectors with the GPT-4 text encoder \cite{gpt4} and ranking them by pairwise cosine similarity; a prediction is a true positive only if it exceeds the $tIOU$ threshold AND its corresponding ground truth description appears in the top-k matches.
Hereafter, we refer to the F1 metric for class-agnostic localization as F1 (loc.), and top-k F1 metric as F1 (loc.+desc.).
Detailed evaluation steps, full GPT-Score results without averaging, and failure cases of N-gram-based metrics are provided in the supplementary materials.

\newcommand\mr[2]{\multicolumn{1}{c}{\multirow{#1}{*}{\makecell{#2}}}}

\begin{table}[t!]
    \centering
    \resizebox{\linewidth}{!}
    {
    \begin{tabular}{c c ccc ccc c}
    \hlineB{2.5}
    \multirow{2}{*}{}                       & \multicolumn{3}{c}{\textbf{F1 (loc.) $\uparrow$}}                                             & \multicolumn{3}{c}{\textbf{F1 (loc. + desc.) $\uparrow$}}                      & \mr{2.87}{\textbf{GPT} \\ \textbf{Score $\uparrow$}}                                     \\ \cmidrule(lr){2-4} \cmidrule(lr){5-7}
                                            & 0.3                           & 0.5                           & 0.7 	   		                & 0.3                           & 0.5                             & 0.7         &                                                               \\ \hlineB{2.5}
    \multicolumn{8}{l}{\cellcolor[gray]{0.9} \textbf{\textit{Annotation}}}                                                                                                                                                                                                                  \\
    EgoGs Pseudo	                        & 64.7                          & 56.1                          & 42.8	                        & 34.04	                        & 30.87	                         & 26.10	    & 2.98                                                          \\ \hlineB{2.5}
    \multicolumn{8}{l}{\cellcolor[gray]{0.9} \textbf{\textit{Train}}}                                                                                                                                                                                                                       \\
    EgoGS GT	                            & 51.58                         & 39.70                         & 24.76	                        & 15.23	                        & 12.67	                          & 8.68		& 2.96                                                          \\
    EgoGS Pseudo	                        & 50.65                         & 38.24                         & 23.48	                        & 14.30	                        & 11.63	                          & 7.66		& 3.02                                                          \\ \hlineB{2.5}
    \end{tabular}}
    \vspace{-3mm}        
    \caption{Evaluation of pseudo-labels from our dataset pipeline. All F1 scores are calculated using the ground truth step labels from the EgoGS dataset.}
    \vspace{-3mm}     
    \label{tab:data-pipeline}
\end{table}
\setlength{\dashlinedash}{0.3pt}  
\setlength{\dashlinegap}{0.3pt}     
\setlength{\arrayrulewidth}{0.3pt} 

\begin{table}[t!]
    \centering
    \resizebox{\linewidth}{!}
    {
    \begin{tabular}{c c c c c}
    \hlineB{2.5}
    \textbf{Dataset}            & \textbf{Annotation}           & \textbf{Hier.}                & \textbf{Hybrid}                 & \textbf{F1 (loc.) $\uparrow$}         \\ \hlineB{2.5}
    \multirow{6}{*}{EgoGS}      & \multirow{4}{*}{GT}           & \multirow{2}{*}{Step}         & \ding{55}                         & 16.78                                 \\
                                &                               &                               & \cellcolor[gray]{0.9} \ding{52}   & \cellcolor[gray]{0.9} \textbf{39.35}  \\ \cdashline{3-5}
                                &                               & \multirow{2}{*}{Substep}      & \ding{55}                         & 22.48                                 \\
                                &                               &                               & \cellcolor[gray]{0.9} \ding{52}   & \cellcolor[gray]{0.9} \textbf{44.79}  \\ \cdashline{2-5}
                                & \multirow{2}{*}{Pseudo}       & \multirow{2}{*}{Step}         & \ding{55}                         & 25.34                                 \\
                                &                               &                               & \cellcolor[gray]{0.9} \ding{52}   & \cellcolor[gray]{0.9} \textbf{37.65}  \\ \hlineB{2.5}
    \multirow{4}{*}{EgEx}       & \multirow{2}{*}{GT}           & \multirow{2}{*}{Substep}      & \ding{55}                         & 8.42                                  \\ 
                                &                               &                               & \cellcolor[gray]{0.9} \ding{52}   & \cellcolor[gray]{0.9} \textbf{52.2}   \\ \cdashline{2-5}
                                & \multirow{2}{*}{Pseudo}       & \multirow{2}{*}{Step}         & \ding{55}                         & 6.12                                  \\
                                &                               &                               & \cellcolor[gray]{0.9} \ding{52}   & \cellcolor[gray]{0.9} \textbf{46.33}  \\ \hlineB{2.5}
    \multirow{4}{*}{EK100}      & \multirow{2}{*}{GT}           & \multirow{2}{*}{Substep}      & \ding{55}                         & 31.46                                 \\ 
                                &                               &                               & \cellcolor[gray]{0.9} \ding{52}   & \cellcolor[gray]{0.9} \textbf{48.95}  \\ \cdashline{2-5}
                                & \multirow{2}{*}{Pseudo}       & \multirow{2}{*}{Step}         & \ding{55}                         & 19.91                                 \\
                                &                               &                               & \cellcolor[gray]{0.9} \ding{52}   & \cellcolor[gray]{0.9} \textbf{34.17}  \\ \hlineB{2.5}
    \end{tabular}}
    \vspace{-3mm}    
    \caption{Ablation study of Hybrid Action Boundary Detection. Since EgEx and EK100 lack ground truth step labels, experiments are conducted using pseudo-step labels.}    
    \vspace{-5mm}
    \label{tab:hybrid-detection}
\end{table}
\newcolumntype{d}[1]{D..{#1}}
\newcommand\mc[1]{\multicolumn{1}{c}{#1}}
\setlength{\dashlinedash}{0.3pt}  
\setlength{\dashlinegap}{0.3pt}     
\setlength{\arrayrulewidth}{0.3pt} 
 
\begin{table*}[t!]
    \centering
    \tiny
    \resizebox{\linewidth}{!}
    {
    \begin{tabular}{>{\centering\arraybackslash}p{1cm} >{\centering\arraybackslash}p{0.5cm} >{\centering\arraybackslash}p{0.5cm} >{\centering\arraybackslash}p{0.5cm} ccc ccc c c} 
    \hlineB{2.5}
    \mr{2.87}{\textbf{Method}} & \mr{2.87}{\textbf{Train}} & \mr{2.87}{\textbf{Valid}} & \mr{2.87}{\textbf{Hier.}} & \multicolumn{3}{c}{\textbf{F1 (loc.) $\uparrow$}} & \multicolumn{3}{c}{\textbf{F1 (loc. + desc.) $\uparrow$}}  & \mr{2.87}{\textbf{GPT} \\ \textbf{Score $\uparrow$}} & \mr{2.87}{\textbf{Goal} \\ \textbf{Acc. $\uparrow$}} \\ \cmidrule(lr){5-7} \cmidrule(lr){8-10}
                                                & 			                            & 				                        &                                       & 0.3                                   & 0.5                                & 0.7 	   	                            & 0.3                                   & 0.5                                   & 0.7                                   &                                        &                                       \\ \hlineB{2.5}
    \multirow{2}{*}{GT Proposal}                & \multirow{2}{*}{-}	                & \multirow{2}{*}{EgoGS}                & Step	                                & -                                     & -    	                             & -    	                            & 28.68	                                & 28.68	                                & 28.68		                            & 2.96                                   & \multirow{2}{*}{47.01}                \\ 	
                                                &                                       &                                       & \cellcolor[gray]{0.9}Substep          & \cellcolor[gray]{0.9}-	            & \cellcolor[gray]{0.9}-    	     & \cellcolor[gray]{0.9}-    	        & \cellcolor[gray]{0.9}33.12	        & \cellcolor[gray]{0.9}33.12	        & \cellcolor[gray]{0.9}33.12	        & \cellcolor[gray]{0.9}2.77              &                                       \\ \hline
    \multirow{2}{*}{TeSTra~\cite{testra}}	    & \multirow{10}{*}{EgoGS}	            & \multirow{14}{*}{EgoGS}	            & Step	                                & 18.27	                                & 12.51	                             & 7.08 	                            & 4.21 	                                & 3.12 	                                & 2.19 		                            & 2.79                                   & \multirow{2}{*}{37.31}                \\ 	
                                                &                                       &                                       & \cellcolor[gray]{0.9}Substep          & \cellcolor[gray]{0.9}22.56	        & \cellcolor[gray]{0.9}15.85	     & \cellcolor[gray]{0.9}9.55            & \cellcolor[gray]{0.9}6.84 	        & \cellcolor[gray]{0.9}4.81 	        & \cellcolor[gray]{0.9}2.72 	        & \cellcolor[gray]{0.9}2.74              &                                       \\ \cdashline{4-12}
    \multirow{2}{*}{MiniROAD~\cite{miniroad}}	&                                       &                                       & Step	                                & 19.23                                 & 12.46	                             & 6.56 	                            & 4.26 	                                & 2.81 	                                & 1.79 		                            & 2.73                                   & \multirow{2}{*}{38.31}                \\ 	
                                                &                                       &                                       & \cellcolor[gray]{0.9}Substep          & \cellcolor[gray]{0.9}32.12	        & \cellcolor[gray]{0.9}21.90	     & \cellcolor[gray]{0.9}12.88           & \cellcolor[gray]{0.9}9.77 	        & \cellcolor[gray]{0.9}7.05 	        & \cellcolor[gray]{0.9}4.14 	        & \cellcolor[gray]{0.9}2.77              &                                       \\ \cdashline{4-12}
    \multirow{2}{*}{MAT~\cite{mat}}	            &                                       &                                       & Step	                                & 18.36                                 & 12.34	                             & 7.00 	                            & 4.01 	                                & 3.01 	                                & 1.98 		                            & 2.82                                   & \multirow{2}{*}{40.30}                \\ 	
                                                &                                       &                                       & \cellcolor[gray]{0.9}Substep          & \cellcolor[gray]{0.9}26.22	        & \cellcolor[gray]{0.9}18.91	     & \cellcolor[gray]{0.9}11.28           & \cellcolor[gray]{0.9}7.32 	        & \cellcolor[gray]{0.9}5.13 	        & \cellcolor[gray]{0.9}3.21 	        & \cellcolor[gray]{0.9}2.74              &                                       \\ \cdashline{1-1}\cdashline{4-12} 
    \multirow{2}{*}{SDVC~\cite{streamingvideocaptioning}}	            &               &                                       & Step	                                & 22.23                                 & 10.43	                             & 3.06 	                            & 2.61 	                                & 1.32 	                                & 0.57 		                            & 2.12                                   & \multirow{2}{*}{-}                    \\ 	
                                                &                                       &                                       & \cellcolor[gray]{0.9}Substep          & \cellcolor[gray]{0.9}17.81	        & \cellcolor[gray]{0.9}7.79	         & \cellcolor[gray]{0.9}2.30            & \cellcolor[gray]{0.9}2.28 	        & \cellcolor[gray]{0.9}0.80 	        & \cellcolor[gray]{0.9}0.22 	        & \cellcolor[gray]{0.9}1.76              &                                       \\ \cdashline{4-12} 
    \multirow{2}{*}{SDVC\textsuperscript{*}~\cite{streamingvideocaptioning}}&           &                                       & Step	                                & 22.23                                 & 10.43	                             & 3.06 	                            & 6.20 	                                & 3.25 	                                & 0.79 		                            & 2.45                                   & \multirow{2}{*}{-}                    \\ 	
                                                &                                       &                                       & \cellcolor[gray]{0.9}Substep          & \cellcolor[gray]{0.9}17.81	        & \cellcolor[gray]{0.9}7.79	         & \cellcolor[gray]{0.9}2.30            & \cellcolor[gray]{0.9}5.73 	        & \cellcolor[gray]{0.9}2.57 	        & \cellcolor[gray]{0.9}0.93 	        & \cellcolor[gray]{0.9}2.18              &                                       \\ \cdashline{1-2} \cdashline{4-12} 
    \multirow{4}{*}{\textbf{OpenHOUSE}}	        & \multirow{2}{*}{EgoGS}                &                                       & Step	                                & \textbf{51.58}                        & \textbf{39.70}	                 & \textbf{24.76}	                    & \textbf{15.23}	                    & \textbf{12.67}	                    & \textbf{8.68} 		                & 2.96                                   & \multirow{2}{*}{\textbf{47.76}}       \\ 	
                                                &                                       &                                       & \cellcolor[gray]{0.9}Substep          & \cellcolor[gray]{0.9}55.17	        & \cellcolor[gray]{0.9}43.71         & \cellcolor[gray]{0.9}28.83           & \cellcolor[gray]{0.9}19.79	        & \cellcolor[gray]{0.9}16.11	        & \cellcolor[gray]{0.9}10.89	        & \cellcolor[gray]{0.9}2.78              &                                       \\ \cdashline{2-2} \cdashline{4-12}
                        	                    & \multirow{2}{*}{EgoGS$^\dagger$}      &                                       & Step	                                & 50.65                                 & 38.24	                             & 23.48	                            & 14.30	                                & 11.63	                                & 7.66 		                            & 3.02                                   & \multirow{2}{*}{45.52}                \\ 	
                                                &                                       &                                       & \cellcolor[gray]{0.9}Substep          & \cellcolor[gray]{0.9}\textbf{55.75}	& \cellcolor[gray]{0.9}\textbf{43.88}& \cellcolor[gray]{0.9}\textbf{29.38}  & \cellcolor[gray]{0.9}\textbf{19.80}	& \cellcolor[gray]{0.9}\textbf{16.42}	& \cellcolor[gray]{0.9}\textbf{11.71}	& \cellcolor[gray]{0.9}2.79              &                                       \\ \hline
    \multirow{6}{*}{\textbf{OpenHOUSE}}         & \multirow{2}{*}{EgEx$^\dagger$}       & \multirow{2}{*}{EgEx$^\dagger$}       & Step	                                & 62.85                                 & 45.75	                             & 26.15	                            & 19.99	                                & 14.76	                                & 9.87 		                            & 3.18                                   & \multirow{2}{*}{90.66}                \\ 	
                                                &                                       &                                       & \cellcolor[gray]{0.9}Substep          & \cellcolor[gray]{0.9}63.42	        & \cellcolor[gray]{0.9}51.12	     & \cellcolor[gray]{0.9}34.36           & \cellcolor[gray]{0.9}23.22	        & \cellcolor[gray]{0.9}19.53	        & \cellcolor[gray]{0.9}14.03	        & \cellcolor[gray]{0.9}2.56              &                                       \\ \cdashline{4-12}
                        	                    & \multirow{2}{*}{EK100$^\dagger$}      & \multirow{2}{*}{EK100$^\dagger$}      & Step	                                & 55.32                                 & 34.48	                             & 15.91	                            & 14.13	                                & 9.04  	                            & 4.35 		                            & 3.24                                   & \multirow{2}{*}{29.71}                \\ 	
                                                &                                       &                                       & \cellcolor[gray]{0.9}Substep          & \cellcolor[gray]{0.9}65.98	        & \cellcolor[gray]{0.9}48.01	     & \cellcolor[gray]{0.9}25.73           & \cellcolor[gray]{0.9}13.54	        & \cellcolor[gray]{0.9}10.51	        & \cellcolor[gray]{0.9}6.09 	        & \cellcolor[gray]{0.9}2.64              &                                       \\ \cdashline{4-12}
                                                & \multirow{2}{*}{Exo.$^\dagger$}	    & \multirow{2}{*}{Exo.$^\dagger$}       & Step	                                & 62.03                                 & 39.73	                             & 19.73	                            & 12.91	                                & 9.18	                                & 4.99                                  & 2.80                                   & \multirow{2}{*}{79.12}                \\ 	
                                                &                                       &                                       & \cellcolor[gray]{0.9}Substep          & \cellcolor[gray]{0.9}57.33	        & \cellcolor[gray]{0.9}40.29	     & \cellcolor[gray]{0.9}21.35           & \cellcolor[gray]{0.9}10.06	        & \cellcolor[gray]{0.9}7.60 	        & \cellcolor[gray]{0.9}4.50             & \cellcolor[gray]{0.9}1.96              &                                       \\ \hlineB{2.5}
    
    \end{tabular}}
    \vspace{-3mm} 
    \caption{Comparison with baseline methods and results in various datasets. $\dagger$ indicates datasets with ``step pseudo-label'' from our pipeline, and ``Exo.'' denotes EgEx's Exo view. ``GT proposal'' refers to experiments using ground truth action proposals as VLM input, without class information. Zero-shot VLM inference achieves around 30\% accuracy in matching ground truth labels. InternVL2-40B-AWQ \cite{internvl2} is used as the frozen VLM. Following the original implementation, YT-Temporal \cite{yt-temporal} pretrained Vid2Seq \cite{vid2seq} weights are used for initialization of SDVC, and fine-tuned using EgoGS annotations, with separate training for each hierarchy. We used the default configuration on SDVC.}
    \label{tab:main-table}
\end{table*} 
\setlength{\dashlinedash}{0.3pt}  
\setlength{\dashlinegap}{0.3pt}     
\setlength{\arrayrulewidth}{0.3pt} 

\begin{table*}[t!]
    \centering
    \tiny
    \resizebox{\linewidth}{!}
    {
    \begin{tabular}{>{\centering\arraybackslash}p{1cm} c c ccc ccc >{\centering\arraybackslash}p{0.7cm}}
    \hlineB{2.5}
    \mr{2.87}{\textbf{Train}}                         & \mr{2.87}{\textbf{Valid}}                   & \mr{2.87}{\textbf{Hier.}}                   & \multicolumn{3}{c}{\textbf{F1 (loc.) $\uparrow$}}                                                                                                 & \multicolumn{3}{c}{\textbf{F1 (loc. + desc.) $\uparrow$}}                                      & \mr{2.87}{\textbf{Goal}\\\textbf{Acc. $\uparrow$}}                 \\ \cmidrule(lr){4-6} \cmidrule{7-9}
    			                                            & 				                                    &                                                   & 0.3                                               & 0.5                                           & 0.7 	   	                                    & 0.3                                           & 0.5                                           & 0.7                                           &                                       \\ \hlineB{2.5}
    \multirow{2}{*}{\color{gray} EK100$^\dagger$}           & \multirow{2}{*}{\color{gray} EK100$^\dagger$}     & \color{gray} Step	                                & \color{gray} 55.32                                & \color{gray} 34.48	                        & \color{gray} 15.91	                        & \color{gray} 14.13	                        & \color{gray} 9.04 	                        & \color{gray} 4.35                             & \multirow{2}{*}{\color{gray} 29.71}   \\
                                                            &                                                   & \cellcolor[gray]{0.9}{\color{gray} Substep}       & \cellcolor[gray]{0.9}{\color{gray} 65.98}	        & \cellcolor[gray]{0.9}{\color{gray} 48.01}	    & \cellcolor[gray]{0.9}{\color{gray} 25.73}	    & \cellcolor[gray]{0.9}{\color{gray} 13.54}	    & \cellcolor[gray]{0.9}{\color{gray} 10.51}	    & \cellcolor[gray]{0.9}{\color{gray} 6.09}      &                                       \\ \hline
    \multirow{2}{*}{EgoGS}                                  & \multirow{6}{*}{EK100$^\dagger$}                  & Step	                                            & 42.49                                             & 25.62	                                        & 12.21	                                        & 11.24	                                        & 7.48 	                                        & 3.98                                          & \multirow{2}{*}{28.26}                \\
                                                            &                                                   & \cellcolor[gray]{0.9}Substep                      & \cellcolor[gray]{0.9}44.91	                    & \cellcolor[gray]{0.9}26.09	                & \cellcolor[gray]{0.9}11.39	                & \cellcolor[gray]{0.9}9.81 	                & \cellcolor[gray]{0.9}6.08 	                & \cellcolor[gray]{0.9}2.72                     &                                       \\ \cdashline{3-10}
    \multirow{2}{*}{EgEx$^\dagger$}                         &                                                   & Step	                                            & 44.39                                             & 26.22	                                        & 10.50	                                        & 11.43	                                        & 7.15 	                                        & 3.29                                          & \multirow{2}{*}{25.36}                \\
                                                            &                                                   & \cellcolor[gray]{0.9}Substep                      & \cellcolor[gray]{0.9}50.75	                    & \cellcolor[gray]{0.9}30.74	                & \cellcolor[gray]{0.9}13.82	                & \cellcolor[gray]{0.9}10.68	                & \cellcolor[gray]{0.9}6.80 	                & \cellcolor[gray]{0.9}3.29                     &                                       \\ \cdashline{3-10}
    \multirow{2}{*}{\textbf{EgoGS + EgEx$^\dagger$}}        &                                                   & \textbf{Step}	                                    & \textbf{51.64}                                    & \textbf{32.47}	                            & \textbf{13.84}	                            & \textbf{12.52}	                            & \textbf{8.84}	                            & \textbf{4.26}                                & \multirow{2}{*}{\textbf{29.71}}            \\
                                                            &                                                   & \cellcolor[gray]{0.9}\textbf{Substep}             & \cellcolor[gray]{0.9}\textbf{53.67}	            & \cellcolor[gray]{0.9}\textbf{33.98}	        & \cellcolor[gray]{0.9}\textbf{15.98}	        & \cellcolor[gray]{0.9}\textbf{11.97}	        & \cellcolor[gray]{0.9}\textbf{8.11}	        & \cellcolor[gray]{0.9}\textbf{4.24}           &                                        \\ \hlineB{2.5}
    \end{tabular}}
    \vspace{-3mm}
    \caption{Cross-validation results. The EK100-trained model serves as the upper bound, while the EgoGS + EgEx-trained model yields comparable results, demonstrating scalability when merging multiple datasets.}
    \vspace{-5mm}
    \label{tab:cross-validation}
\end{table*}
\subsection{Results and Analysis}
\label{sec:results_and_analysis}
\noindent\textbf{Validating Dataset Pipeline}
Table \ref{tab:data-pipeline} presents the validation of our dataset pipeline. 
Since the Ego4D-Goalstep dataset contains both substep and step annotations, we compare the generated pseudo step labels from our pipeline with the ground truth step labels. 
The results in the first row show that the LLM effectively groups atomic actions, leading to a reasonable approximation of the original step annotations. 
However, due to the inherent subjectivity in action temporal boundaries, perfect recovery is obviously impossible.

More importantly, we assess whether these pseudo-labels provide \textit{meaningful learning signals} for step-level hierarchy training.
Surprisingly, models trained with our pseudo-labels performed competitively with those trained on the actual ground truth labels. 
This suggests that even though the generated step labels do not exactly match the originals, they do not contain critical errors but instead provide a plausible alternative. 
Just as different human annotators might produce varying temporal boundaries, the generated labels still offer a valid interpretation rather than a completely incorrect one.

\noindent\textbf{Hybrid Action Boundary Detection}
Table \ref{tab:hybrid-detection} demonstrates the effectiveness of our hybrid action boundary detection method. 
Across all configurations and datasets, incorporating our hybrid method yields significant gains.
It highlights the critical importance of our approach in processing procedural videos, where many instances are strictly adjacent.

\noindent\textbf{Comparison with Other Methods}
As the first work in Hierarchical Streaming Video Understanding, we establish two baseline approaches:
(i) We extend recent OAD methods by training class-agnostic variants of \cite{testra, miniroad, mat} using temporal annotations. For detected action instances, we apply the same description generation process as OpenHOUSE.
(ii) We compare against SDVC \cite{streamingvideocaptioning}, treating the task as an extension of streaming dense video captioning.
We introduce two SDVC variants: fine-tuned one (SDVC) following the original implementation and another (SDVC*) that uses only the action proposals from the fine-tuned model while following OpenHOUSE’s description generation process with a frozen large-scale VLM.

Table \ref{tab:main-table} shows OpenHOUSE significantly outperforms both baselines across all metrics, yet comparing the baselines provides additional insights.
The higher F1 (loc.) of the OAD extension (TeSTra, MiniROAD, MAT) over SDVC suggests OAD-based approaches are more effective for online localization, supporting our choice of extending OAD-based On-TAL in OpenHOUSE.
Additionally, the better performance of SDVC* over SDVC indicates that, for relatively small-scale datasets, leveraging a powerful pre-trained VLM may be more effective than full fine-tuning.

Moreover, the goal accuracy gap between baselines and our method demonstrates that accurate lower-hierarchy action predictions are crucial for identifying high-level semantics, as they provide relevant information. 
Additionally, experiments on the exo-view in EgoExo4D show that OpenHOUSE is not limited to egocentric views, though third-person view is not the primary focus of this paper.

\noindent\textbf{Cross-Dataset Evaluation}
Thanks to ``class-agnostic'' streaming module and VLM's ``zero-shot'' inference strategy, OpenHOUSE can be trained on a merged dataset and evaluated across different datasets.

For the experiment, we fixed the validation dataset to EK100 while varying the training datasets. 
Table \ref{tab:cross-validation} shows the cross-dataset evaluation results.
Comparing the EK100-trained configuration with others, OpenHOUSE demonstrates strong performance even in cross-dataset scenarios. 
Notably, we observe scalability: training on the merged dataset (EgoGS + EgoEx) yields the highest F1 score.
This indicates that OpenHOUSE can be further extended with more training videos, highlighting the strong scalability and potential of the framework.
\setlength{\dashlinedash}{0.3pt}  
\setlength{\dashlinegap}{0.3pt}     
\setlength{\arrayrulewidth}{0.3pt} 
\renewcommand{\arraystretch}{1.2}

\begin{table}[t!]
    \centering
    \scriptsize
    \resizebox{\linewidth}{!}
    {
    
    \begin{tabular}{>{\centering\arraybackslash}m{1.2cm} >{\centering\arraybackslash}m{1cm}   c c c    c    c c c c  >{\centering\arraybackslash}m{0.6cm}}
    \hlineB{2.5}
    \mr{2.87}{\textbf{}}                                                & \mr{2.87}{\textbf{Hier.}}               & \multicolumn{2}{c}{\textbf{SDVC~\cite{streamingvideocaptioning}}}       & \mr{2.87}{\textbf{OpenHOUSE}}          \\ \cmidrule(lr){3-4} 
                                                                        &                                         & 70                                    & 105                                &                                   \\ \hlineB{2.5}
    \mr{2}{\textbf{AEDT (s)} $\downarrow$}	                                & Step	                                  & 36.36                                 & 64.84  		                    & 4.94                                \\
                                                                        & \cellcolor[gray]{0.9}Substep            & \cellcolor[gray]{0.9}43.95            & \cellcolor[gray]{0.9}52.64 		& \cellcolor[gray]{0.9}1.82         \\ \cdashline{2-6}
    \mr{2}{\textbf{F1@0.3} \\ \textbf{(loc.) $\uparrow$}}            & Step	                                        & 9.85		                            & 14.12                              & 51.58       		                 \\ 
                                                                        & \cellcolor[gray]{0.9}Substep            & \cellcolor[gray]{0.9}5.48		    & \cellcolor[gray]{0.9}8.55        & \cellcolor[gray]{0.9}55.17	     \\ \hlineB{2.5}
    \end{tabular}}
    \vspace{-3mm}
    \caption{Comparison of prediction delay (AEDT~\cite{oat}) and F1 (loc.) between SDVC~\cite{streamingvideocaptioning} and OpenHOUSE in EgoGS~\cite{ego4dgoalstep}. The numbers 70 and 105 denote the decoding points interval of SDVC.}    
    \vspace{-3mm}
    \label{tab:aedt}
\end{table}
\begin{table}[t!]
    \centering
    \scriptsize
    \resizebox{\linewidth}{!}
    {
    \begin{tabular}{c c c c c c ccc}
    \hlineB{2.5}
    \multicolumn{1}{c}{\textbf{Num.}}         & \multicolumn{1}{c}{\textbf{Frames}}           & \multicolumn{1}{c}{\textbf{Previous}}     & \multicolumn{1}{c}{\textbf{F1}}     & \multicolumn{1}{c}{\textbf{GPT}}             \\ 
    \textbf{Frames}                           & \textbf{Hier.}                                & \textbf{Output}                           & \textbf{(loc. + desc.)$\uparrow$}    & \textbf{Score}            	                    \\ \hlineB{2.5}
    59385                                     & \ding{55}                                     & \ding{55}                                 & 13.08	            	            & 2.59                                              \\  	
    22770                                     & \ding{52}                                     & \ding{55}                                 & 10.96             	                & 2.42                                               \\  	
    22770                                     & \ding{52}                                     & \ding{52}                                 & 13.05             	                & 2.76                                               \\ \hlineB{2.5} 	
    \end{tabular}}
    \vspace{-3mm}     
    \caption{Ablation study on different uses of context memory for generating EgoGS step captions. ``Frames Hier." refers to the hierarchical-aware frame sampling strategy, while ``Previous Output" refers to using previous predictions as auxiliary information. InternVL2-76B~\cite{internvl2} is used across all the experiments.}
    \vspace{-7mm}    
    \label{tab:context-memory}
\end{table}

\begin{figure}
    \centering
    \includegraphics[width=\linewidth]{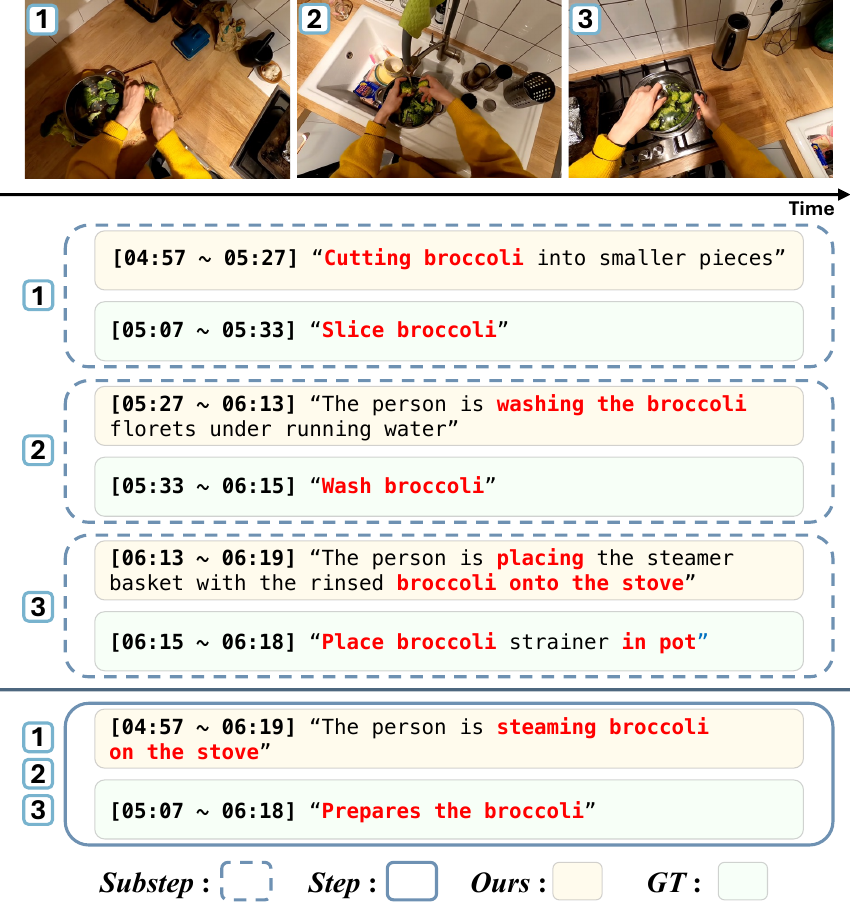}
    \caption{Qualitative results. For the GT substep caption \textit{"Place broccoli strainer in pot"}, OpenHOUSE provides \textit{"The person is placing the steamer basket with the rinsed broccoli onto the stove."}, which aligns better with the actual visual content.}
    \label{fig:qualitative-result}
    \vspace{-6mm}
\end{figure}

\noindent\textbf{Prediction Delay}
To quantitatively assess the promptness of OpenHOUSE, we compare its prediction delay with SDVC using the AEDT metric from \cite{oat}, which measures the time gap between when a true positive prediction is generated and when it should ideally be produced.
Table \ref{tab:aedt} presents the results, showing that SDVC~\cite{streamingvideocaptioning} introduces significant delays compared to our strictly online approach.
Moreover, simply shortening the interval between decoding points significantly degrades localization performance, as noted in \cite{streamingvideocaptioning}, making it unsuitable for online approaches.

\noindent\textbf{Utilization of Context Memory}
As discussed in Section \ref{sec:vlm}, we leverage context memory during VLM inference.
The key information in the context memory includes hierarchical frame membership (e.g., used when selecting frames within the substeps of the current step for VLM inference) and previous text-based predictions. 
Table \ref{tab:context-memory} presents the results of these experiments, showing that incorporating context memory that leverages the hierarchical structure improves performance while requiring only 40\% of the frames. 
This reduction in frame usage significantly lowers the computational complexity of VLM inference.

\noindent\textbf{Inference speed}
We evaluated the inference efficiency on a 46-minute video at 16 fps (totaling 2758 * 16 frames) using four RTX 3090 GPUs with the VLM model InternVL2-40B-AWQ~\cite{internvl2}. OpenHOUSE achieves an average processing speed of 24 fps, which is \textbf{16 times faster} compared to using the VLM for every frame in the same environment. This shows that our framework significantly improves the efficiency of VLM utilization given the same input.

\noindent\textbf{Qualitative results}
Figure \ref{fig:qualitative-result} shows the qualitative results of our OpenHOUSE framework. 
The video is processed in an online manner, generating both substep and step captions.
We observed that the generated captions exhibit high semantic similarity to the ground truth (GT) captions, providing even more detailed descriptions.

\noindent\textbf{Additional experiments} Experiments on various VLMs, description generation for incomplete action instances, additional details on the mentioned experiments, and demo videos are provided in the supplementary materials.

\section{Conclusion}
With advancing VLMs, we see their integration into streaming settings as the next step for online action perception.
In this context, OpenHOUSE is a key step, laying the foundation for future research.
\clearpage
\setcounter{page}{9}
\maketitlesupplementary
\appendix

\begin{strip}
    \centering
    \includegraphics[width=\linewidth]{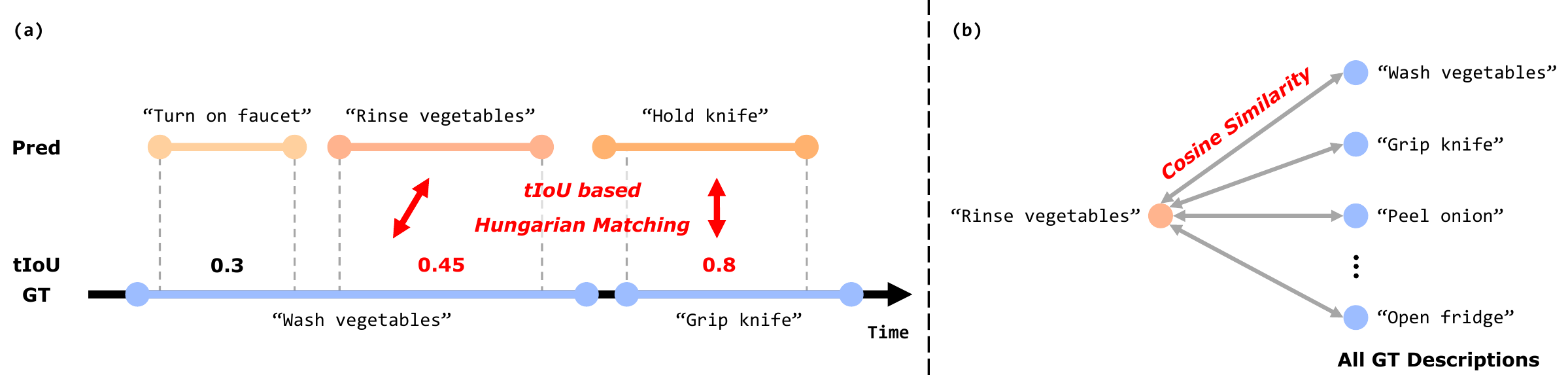}
    \captionof{figure}{(a) $tIoU$-based Hungarian matching.
    Hungarian matching provides optimal one-to-one matching in terms of maximizing tIoU, unlike greedy matching which allows duplication. 
    Here, the algorithm selects \textit{"Rinse vegetables"} as the best match for the GT \textit{"Wash vegetables"}. 
    With a predefined tIoU threshold of 0.3, \textit{"Rinse vegetables"} is considered a True Positive (TP), while the unmatched prediction \textit{"Turn on faucet"} is classified as a False Positive (FP). If no prediction matches the GT or fails to meet the threshold, the instance is counted as a False Negative (FN).
    (b) The description \textit{"Rinse vegetables"}, initially marked as a TP in F1 (loc.), is encoded into an embedding vector. Similarly, all GT descriptions in the test split are encoded, and pairwise cosine similarities are computed. If \textit{"Wash vegetables"} appears among the top 5 most similar GT descriptions, \textit{"Rinse vegetables"} is confirmed as a TP.
    }
    \label{fig:metric_1}
\end{strip}

\section{Details of Metrics}
\label{sec:metric}

This section provides detailed explanations of the evaluation metrics presented in the main text.

\begin{algorithm}
\caption{\texttt{get\_hungarian\_score} in Python.}
\definecolor{codeblue}{rgb}{0.25,0.5,0.5}
\lstset{
  backgroundcolor=\color{white},
  basicstyle=\fontsize{7.5pt}{7.5pt}\ttfamily\selectfont,
  columns=fullflexible,
  breaklines=true,
  captionpos=b,
  commentstyle=\fontsize{7.2pt}{7.2pt}\color{codeblue},
  keywordstyle=\fontsize{7.2pt}{7.2pt},
  texcl=<true|false>
}
\begin{lstlisting}[language=python]
def get_hungarian_score(answer:list, prediction:list, iou_threshold=0.5):
    '''
    IN: answer[[st,ed],[st,ed]...],
    prediction[[st,ed],[st,ed]...]
    OUT: f1_score (float)
    '''
    profit = np.zeros((len(answer), len(prediction)))
    #calculate pairwise iou
    for i in range(len(answer)):
        for j in range(len(prediction)):
            profit[i][j] = calculate_iou(answer[i], prediction[j])
    #perform Hungarian Matching
    r, c = optimize.linear_sum_assignment(profit, maximize=True)
    tp = np.sum(np.where(profit[r, c] >= iou_threshold, 1, 0))
    a = answer.shape[0]
    p = prediction.shape[0]
    #return F1 score
    return 2*tp/(a+p)
\end{lstlisting}
\label{alg:hungarian}
\end{algorithm}

\noindent\textbf{Class-agnostic F1 [\textit{F1 (loc.)}]}
Previous work \cite{actionswitch} conducted an in-depth study on appropriate metrics for evaluating class-agnostic action proposals in a streaming setting, revealing that Hungarian F1 is an effective measure of performance. 
Accordingly, we use Hungarian F1 (hereafter, \textit{F1 (loc.)}) to evaluate the streaming perception module.
This metric employs the Hungarian matching algorithm \cite{hungarian}, which provides optimal bipartite matching between class-agnostic ground truth $\{(t^s_m, t^e_m)\}_{m=1}^M$ and the model's predictions $\{(\hat{t}^s_m, \hat{t}^e_m)\}_{m=1}^{\hat{M}}$ in terms of $tIoU$.
A prediction is considered a true positive if the overlap with the matched GT exceeds a predefined $tIoU$ (in our paper, $tIoU \in \{0.3,0.5,0.7\}$) threshold.
Figure \ref{fig:metric_1} (a) illustrates tIoU-based Hungarian matching and Algorithm \ref{alg:hungarian} presents a brief Python implementation for calculating \textit{F1 (loc.)}.

\begin{figure*}
    \centering
    \includegraphics[width=\linewidth]{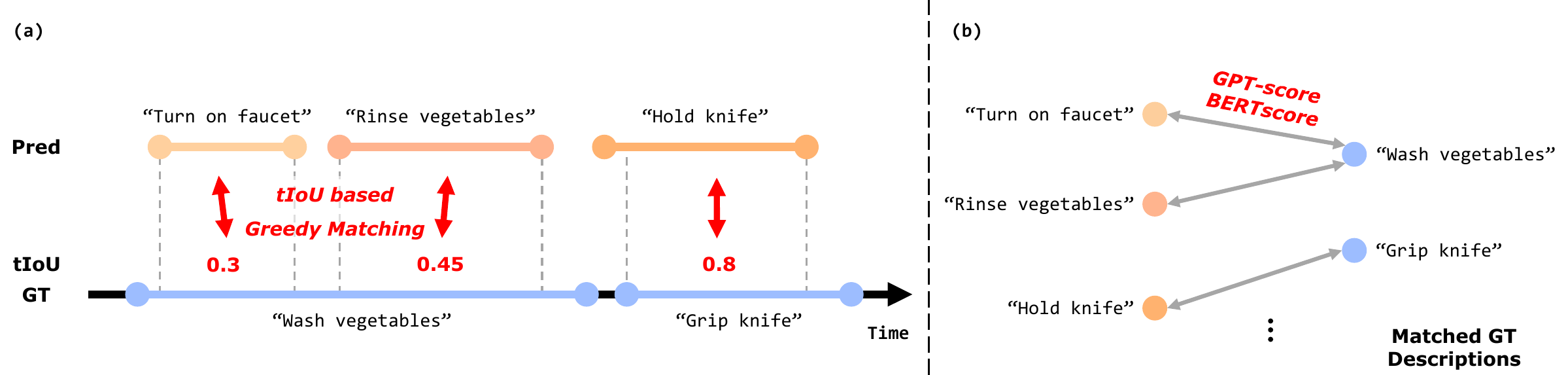}
    \caption{An example of calculating GPT, or BERTscore.
    With a $tIoU$ threshold of 0.3, the predictions \textit{"Turn on faucet"} and \textit{"Rinse vegetables"} are matched with the GT \textit{"Wash vegetables"}, while \textit{"Hold knife"} is matched with \textit{"Grip knife"}. 
    This duplicated matching with \textit{"Wash vegetables"} highlights a key difference from the Hungarian F1 metric.
    After matching, GPT-Score or BERTscore is computed for each matched pair, and the final score is obtained by averaging all pair scores.}   
    \label{fig:metric_2}
\end{figure*}

\noindent\textbf{Top-k F1 [\textit{F1 (loc. + desc.)}]}
While \textit{F1 (loc.)} only measures class-agnostic action proposals, our final output includes free-form descriptions of action instances.
To account for this, we extend \textit{F1 (loc.)} to define \textit{F1 (loc. + desc.)}, adding a semantic relevance constraint for True Positives (TP) beyond the $tIoU$ condition.
First, tIoU-based Hungarian matching is performed, identical to \textit{F1 (loc.)}.
Predictions that pass the $tIoU$ threshold are considered candidates. 
Each candidate prediction $\{s_p,e_p,d_p\}$ is matched to a ground truth (GT) instance $\{s_{gt},e_{gt},d_{gt}\}$ where $s$ and $e$ are the start and end times, and $d$ represents the description. (Section \ref{sec:problem_def})

\begin{figure}
    \centering
    \includegraphics[width=\linewidth]{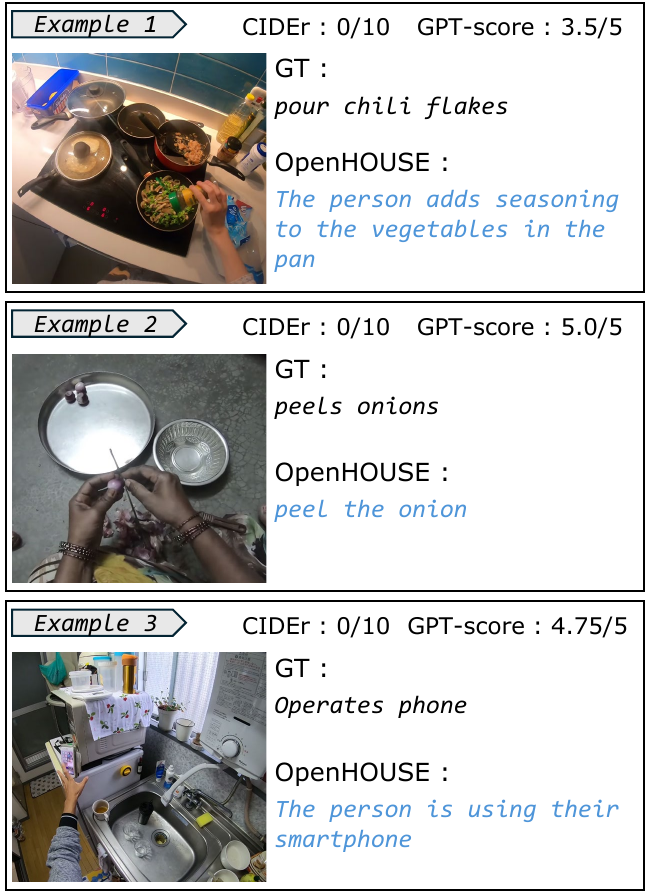}
    \caption{Examples of CIDEr~\cite{cider} and GPT-Score~\cite{videochatgpt} results from EgoGS dataset. Each score was computed at tIoU 0.3. GPT-Score reflects the average of four semantic dimensions: CU, CO, DO, and TU.}
    \label{fig:CIDEr_bad}
    \vspace{-6mm}
\end{figure}
\newcolumntype{d}[1]{D..{#1}}
\setlength{\dashlinedash}{0.3pt}  
\setlength{\dashlinegap}{0.3pt}     
\setlength{\arrayrulewidth}{0.3pt} 

\begin{table}[t!]
    \centering
    \tiny
    \resizebox{\linewidth}{!}
    {
    \begin{tabular}{c c c c}
    \hlineB{2.5}
    \textbf{Method}                 & \textbf{Heir.}                    & \textbf{CIDEr}                & \textbf{METEOR}               \\ \hlineB{2.5}
    \multirow{2}{*}{GT Proposal}    & Step                              & 17.7739                       & 8.4363                        \\  	
                                    & \cellcolor[gray]{0.9}Substep      & \cellcolor[gray]{0.9}32.8570  & \cellcolor[gray]{0.9}10.7680  \\ \hline 
    \multirow{2}{*}{OpenHOUSE}      & Step                              & 7.1885                        & 4.1640                        \\  	
                                    & \cellcolor[gray]{0.9}Substep      & \cellcolor[gray]{0.9}14.3946  & \cellcolor[gray]{0.9}5.5151   \\ \hlineB{2.5}
    \end{tabular}}
    \vspace{-3mm}     
    \caption{CIDEr, METEOR in EgoGS using evaluation tool~\cite{krishna2017dense}.}   
    \label{tab:ngram-based}
    \vspace{-6mm}
\end{table}

The final step is to determine whether $d_p$ and $d_{gt}$ match or not.
Since $d_p$ is a free-form output from a powerful large-scale VLM, it tends to be highly detailed, making semantic relevance crucial—something traditional N-gram-based metrics fail to capture.
To effectively evaluate the semantic relevance of the matched prediction, we use a zero-shot evaluation approach inspired by CLIP \cite{clip}. 
For each matched prediction, we encode the candidate description $d_p$ and all GT descriptions $\{d_m\}_{m=1}^M$ using a GPT-4 \cite{gpt4} text encoder, where $M$ represents the total number of GT descriptions in the whole test split. 
Pairwise cosine similarity is then calculated between $d_p$ and all GT descriptions, which are subsequently ranked by similarity. 
If the matched $d_{gt}$ appears in the top-k rankings, the prediction is finally classified as a TP.
Figure \ref{fig:metric_1} (b) provides an example of this process.
For all experiment, we choose $k=5$ as we empirically found that at least five ground truth descriptions often share near-identical semantics.

\newcounter{promptcounter}
\renewcommand{\thepromptcounter}{\arabic{promptcounter}}

\newenvironment{prompt}[1][htb]
  {\floatname{algorithm}{Prompt}
   \refstepcounter{promptcounter}
   \renewcommand{\thealgorithm}{\thepromptcounter}
   \begin{algorithm}[#1]}
  {\end{algorithm}}

\begin{prompt}
\caption{\texttt{CI Prompt} for Evaluation.}
\definecolor{codeblue}{rgb}{0.25,0.5,0.5}
\lstset{
  backgroundcolor=\color{white},
  basicstyle=\fontsize{6.5pt}{6.5pt}\ttfamily\selectfont,
  columns=fullflexible,
  breaklines=true,
  captionpos=b,
  commentstyle=\fontsize{6.2pt}{6.2pt}\color{codeblue},
  keywordstyle=\fontsize{6.2pt}{6.2pt},
  texcl=<true|false>
}
\begin{lstlisting}
role: system,
content: 
    You are an intelligent chatbot designed for evaluating the factual accuracy of generative outputs for video-based question-answer pairs. 
    Your task is to compare the predicted answer with the correct answer and determine if they are factually consistent. Here's how you can accomplish the task:
    ------
    ##INSTRUCTIONS: 
    - Focus on the factual consistency between the predicted answer and the correct answer. The predicted answer should not contain any misinterpretations or misinformation.
    - The predicted answer must be factually accurate and align with the video content.
    - Consider synonyms or paraphrases as valid matches.
    - Evaluate the factual accuracy of the prediction compared to the answer.

role: user,
content:
    Please evaluate the following video-based question-answer pair:"
    Question: {question}
    Correct Answer: {answer}
    Predicted Answer: {pred}
    Provide your evaluation only as a factual accuracy score where the factual accuracy score is an integer value between 0 and 5, with 5 indicating the highest level of factual consistency. 
    Please generate the response in the form of a Python dictionary string with keys 'score', where its value is the factual accuracy score in INTEGER, not STRING.
    DO NOT PROVIDE ANY OTHER OUTPUT TEXT OR EXPLANATION. Only provide the Python dictionary string. 
    For example, your response should look like this: {''score': 4.8}.
\end{lstlisting}
\vspace{3mm}
\label{prompt:CI}
\end{prompt}

\begin{prompt}
\caption{\texttt{DO Prompt} for Evaluation.}
\definecolor{codeblue}{rgb}{0.25,0.5,0.5}
\lstset{
  backgroundcolor=\color{white},
  basicstyle=\fontsize{6.5pt}{6.5pt}\ttfamily\selectfont,
  columns=fullflexible,
  breaklines=true,
  captionpos=b,
  commentstyle=\fontsize{6.2pt}{6.2pt}\color{codeblue},
  keywordstyle=\fontsize{6.2pt}{6.2pt},
  texcl=<true|false>
}
\begin{lstlisting}
role: system,
content:
    You are an intelligent chatbot designed for evaluating the detail orientation of generative outputs for video-based question-answer pairs. 
    Your task is to compare the predicted answer with the correct answer and determine its level of detail, considering both completeness and specificity. Here's how you can accomplish the task:
    ------
    ##INSTRUCTIONS: 
    - Check if the predicted answer covers all major points from the video. The response should not leave out any key aspects.
    - Evaluate whether the predicted answer includes specific details rather than just generic points. It should provide comprehensive information that is tied to specific elements of the video.
    - Consider synonyms or paraphrases as valid matches.
    - Provide a single evaluation score that reflects the level of detail orientation of the prediction, considering both completeness and specificity.

role: user,
content:
    Please evaluate the following video-based question-answer pair:
    Question: {question}
    Correct Answer: {answer}
    Predicted Answer: {pred}
    Provide your evaluation only as a detail orientation score where the detail orientation score is an integer value between 0 and 5, with 5 indicating the highest level of detail orientation. 
    Please generate the response in the form of a Python dictionary string with keys 'score', where its value is the detail orientation score in INTEGER, not STRING.
    DO NOT PROVIDE ANY OTHER OUTPUT TEXT OR EXPLANATION. Only provide the Python dictionary string. 
    For example, your response should look like this: {''score': 4.8}.
\end{lstlisting}
\label{prompt:DO}
\end{prompt}

\begin{prompt}
\caption{\texttt{CU Prompt} for Evaluation.}
\definecolor{codeblue}{rgb}{0.25,0.5,0.5}
\lstset{
  backgroundcolor=\color{white},
  basicstyle=\fontsize{6.5pt}{6.5pt}\ttfamily\selectfont,
  columns=fullflexible,
  breaklines=true,
  captionpos=b,
  commentstyle=\fontsize{6.2pt}{6.2pt}\color{codeblue},
  keywordstyle=\fontsize{6.2pt}{6.2pt},
  texcl=<true|false>
}
\begin{lstlisting}
role: system,
content:
    You are an intelligent chatbot designed for evaluating the contextual understanding of generative outputs for video-based question-answer pairs. 
    Your task is to compare the predicted answer with the correct answer and determine if the generated response aligns with the overall context of the video content. Here's how you can accomplish the task:
    ------
    ##INSTRUCTIONS: 
    - Evaluate whether the predicted answer aligns with the overall context of the video content. It should not provide information that is out of context or misaligned.
    - The predicted answer must capture the main themes and sentiments of the video.
    - Consider synonyms or paraphrases as valid matches.
    - Provide your evaluation of the contextual understanding of the prediction compared to the answer.

role: user,
content:
    Please evaluate the following video-based question-answer pair:
    Question: {question}
    Correct Answer: {answer}
    Predicted Answer: {pred}
    Provide your evaluation only as a contextual understanding score where the contextual understanding score is an integer value between 0 and 5, with 5 indicating the highest level of contextual understanding. 
    Please generate the response in the form of a Python dictionary string with keys 'score', where its value is contextual understanding score in INTEGER, not STRING.
    DO NOT PROVIDE ANY OTHER OUTPUT TEXT OR EXPLANATION. Only provide the Python dictionary string. 
    For example, your response should look like this: {''score': 4.8}.
\end{lstlisting}
\label{prompt:CU}
\end{prompt}

\begin{prompt}
\caption{\texttt{TU Prompt} for Evaluation.}
\definecolor{codeblue}{rgb}{0.25,0.5,0.5}
\lstset{
  backgroundcolor=\color{white},
  basicstyle=\fontsize{6.5pt}{6.5pt}\ttfamily\selectfont,
  columns=fullflexible,
  breaklines=true,
  captionpos=b,
  commentstyle=\fontsize{6.2pt}{6.2pt}\color{codeblue},
  keywordstyle=\fontsize{6.2pt}{6.2pt},
  texcl=<true|false>
}
\begin{lstlisting}
role: system,
content:
    You are an intelligent chatbot designed for evaluating the temporal understanding of generative outputs for video-based question-answer pairs. 
    Your task is to compare the predicted answer with the correct answer and determine if they correctly reflect the temporal sequence of events in the video content. Here's how you can accomplish the task:
    ------
    ##INSTRUCTIONS: 
    - Focus on the temporal consistency between the predicted answer and the correct answer. The predicted answer should correctly reflect the sequence of events or details as they are presented in the video content.
    - Consider synonyms or paraphrases as valid matches, but only if the temporal order is maintained.
    - Evaluate the temporal accuracy of the prediction compared to the answer.

role: user,
content:
    Please evaluate the following video-based question-answer pair:
    Question: {question}
    Correct Answer: {answer}
    Predicted Answer: {pred}
    Provide your evaluation only as a temporal accuracy score where the temporal accuracy score is an integer value between 0 and 5, with 5 indicating the highest level of temporal consistency. 
    Please generate the response in the form of a Python dictionary string with keys 'score', where its value is the temporal accuracy score in INTEGER, not STRING.
    DO NOT PROVIDE ANY OTHER OUTPUT TEXT OR EXPLANATION. Only provide the Python dictionary string. 
    For example, your response should look like this: {''score': 4.8}.
\end{lstlisting}
\vspace{1mm}
\label{prompt:TU}
\end{prompt}

\noindent\textbf{Description Quality Metrics}
In addition to the main metrics, we use another metrics to evaluate the generated descriptions: GPT-Score~\cite{videochatgpt} and BERTscore~\cite{bertscore}, as introduced in Section \ref{sec:experiment} of the main text.  
For GPTScore, we follow \cite{videochatgpt}, scoring all true positive predictions on a 1–5 scale across the CI (Correctness of Information), DO (Detailed Orientation), CU (Contextual Understanding), and TU (Temporal Understanding) categories using GPT-3.5. The CO (Consistency) category is excluded as it is not suitable for evaluating hierarchical action descriptions. Detailed prompts used for evaluating the CI, DO, CU, and TU categories can be found in Prompt \ref{prompt:CI}, \ref{prompt:DO}, \ref{prompt:CU}, \ref{prompt:TU}.
Unlike \textit{F1 (loc.)}, which identifies optimal one-to-one matching between ground truth and predictions, these metrics use \textit{greedy matching}, allowing multiple predictions to match the same ground truth. A predefined $tIoU$ threshold is applied to filter predictions before calculating the scores.

\noindent\textbf{N-gram based metric}
We found that popular N-gram-based methods (e.g. CIDEr \cite{cider}) are not suitable for evaluating the caption quality of OpenHOUSE. 
Since we use a powerful VLM in a zero-shot setting, it often produces detailed descriptions that may align even better with the actual action than the ground truth, but may not share the exact 
\clearpage
vocabulary with the ground truth. 
This is not a flaw to penalize but rather aligns with our ultimate goal of open-ended understanding.
However, N-gram-based methods require exact word matches, which unjustly penalizes our more detailed descriptions.

Figure \ref{fig:CIDEr_bad} illustrates some examples of this.
Unlike N-gram based method (CIDEr \cite{cider}), we found that the metrics that leverage large-scale language models, such as GPT-based scoring (GPT-Score \cite{videochatgpt}), offers a more robust evaluation framework for semantic similarity.
Note that our main metric, \textit{F1 (loc. + desc.)}, also uses a powerful VLM's text encoder \cite{gpt4} for encoding the descriptions into vector representations, effectively capturing semantic correspondence.
For completeness, we provide results of N-gram based metric in Table \ref{tab:ngram-based}.

\section{Utilization of Different VLMs}
\label{sec:util_diff_vlm}
One of the key components of our system is the VLM, which generates free-form descriptions.
Table \ref{tab:different-vlm} shows the variation in caption quality with different VLMs.
The results indicate that while more advanced VLMs generally improve performance, there is a point of diminishing gains once the VLM is sufficiently powerful. 
This highlights the inherent complexity of Hierarchical Streaming Video Understanding; further improvements cannot rely solely on VLM advancements—enhancing the streaming module is also crucial for meaningful progress.
\setlength{\dashlinedash}{0.3pt}  
\setlength{\dashlinegap}{0.3pt}     
\setlength{\arrayrulewidth}{0.3pt} 

\begin{table}[b]
    \centering
    \vspace{-3mm}      
    \resizebox{\linewidth}{!}
    {
    \begin{tabular}{>{\centering\arraybackslash}p{1cm}cccccc}
    \hlineB{2.5}
    \mr{2.87}{\textbf{Method}}              & \mr{2.87}{\textbf{Completion}}      & \mr{2.87}{\textbf{Hier.}}             & \multicolumn{3}{c}{\textbf{F1 (loc. + desc.) $\uparrow$}}                                                                                  & \mr{2.87}{\textbf{Goal}\\\textbf{Acc. $\uparrow$}}    \\ \cmidrule(lr){4-6}
                                            &                                   &                                       & 0.3                                               & 0.5                                           & 0.7 	   	                            &                                                       \\ \hlineB{2.5}
    \multirow{8}{*}{GT}                     & \multirow{2}{*}{25\%}             & Step                                  &                       18.93                       &                       18.93                   &                       18.93           & \multirow{2}{*}{39.55}                                \\
                                            &                                   & \cellcolor[gray]{0.9}Substep          & \cellcolor[gray]{0.9} 23.66                       & \cellcolor[gray]{0.9} 23.66                   & \cellcolor[gray]{0.9} 23.66           &                                                       \\ \cdashline{3-7}
                                            & \multirow{2}{*}{50\%}             & Step                                  &                       24.56                       &                       24.56                   &                       24.56           & \multirow{2}{*}{45.52}                                \\
                                            &                                   & \cellcolor[gray]{0.9}Substep          & \cellcolor[gray]{0.9} 28.52                       & \cellcolor[gray]{0.9} 28.52                   & \cellcolor[gray]{0.9} 28.52           &                                                       \\ \cdashline{3-7}
                                            & \multirow{2}{*}{75\%}             & Step                                  &                       26.53                       &                       26.53                   &                       26.53           & \multirow{2}{*}{47.76}                                \\
                                            &                                   & \cellcolor[gray]{0.9}Substep          & \cellcolor[gray]{0.9} 31.78                       & \cellcolor[gray]{0.9} 31.78                   & \cellcolor[gray]{0.9} 31.78           &                                                       \\ \cdashline{3-7}
                                            & \multirow{2}{*}{100\%}            & Step                                  &                       28.68                       &                       28.68                   &                       28.68           & \multirow{2}{*}{47.01}                                \\
                                            &                                   & \cellcolor[gray]{0.9}Substep          & \cellcolor[gray]{0.9} 33.12                       & \cellcolor[gray]{0.9} 33.12                   & \cellcolor[gray]{0.9} 33.12           &                                                       \\ \hline
    \multirow{8}{*}{OpenHouse}              & \multirow{2}{*}{25\%}             & Step                                  &                       9.68                       &                       7.86                   &                       5.36           & \multirow{2}{*}{44.03}                                \\
                                            &                                   & \cellcolor[gray]{0.9}Substep          & \cellcolor[gray]{0.9} 13.55                       & \cellcolor[gray]{0.9} 11.14                   & \cellcolor[gray]{0.9} 8.05           &                                                       \\ \cdashline{3-7}
                                            & \multirow{2}{*}{50\%}             & Step                                  &                       12.84                       &                       10.66                   &                       7.45           & \multirow{2}{*}{42.54}                                    \\
                                            &                                   & \cellcolor[gray]{0.9}Substep          & \cellcolor[gray]{0.9} 16.99                       & \cellcolor[gray]{0.9} 13.93                   & \cellcolor[gray]{0.9} 9.91           &                                                       \\ \cdashline{3-7}
                                            & \multirow{2}{*}{75\%}             & Step                                  &                       14.36                       &                       12.02                   &                       8.38           & \multirow{2}{*}{45.52}                                    \\
                                            &                                   & \cellcolor[gray]{0.9}Substep          & \cellcolor[gray]{0.9} 18.8                       & \cellcolor[gray]{0.9} 15.41                   & \cellcolor[gray]{0.9} 10.66           &                                                       \\ \cdashline{3-7}
                                            & \multirow{2}{*}{100\%}            & Step                                  &                       15.23                       &                       12.67                   &                       8.680           & \multirow{2}{*}{47.76}                                \\
                                            &                                   & \cellcolor[gray]{0.9}Substep          & \cellcolor[gray]{0.9} 19.79                       & \cellcolor[gray]{0.9} 16.11                   & \cellcolor[gray]{0.9} 10.89           &                                                       \\ \hlineB{2.5}
    \end{tabular}}
    \vspace{-3mm}      
    \caption{Experiments on EgoGS validating description generation for incomplete action instances.}  
    \label{tab:supp_B}
\end{table}
\begin{table*}[t!]
    \centering
    \scriptsize
    \resizebox{\linewidth}{!}
    {
    
    \begin{tabular}{>{\centering\arraybackslash}p{2.4cm} >{\centering\arraybackslash}p{0.6cm}   c c c    c    c c c c  >{\centering\arraybackslash}p{0.6cm}}
    \hlineB{2.5}
    \mr{2.87}{\textbf{Model}}               & \mr{2.87}{\textbf{Hier.}}               & \multicolumn{3}{c}{\textbf{F1 (loc. + cap.) $\uparrow$}}                                                    & \multicolumn{4}{c}{\textbf{GPT-Score $\uparrow$}~\cite{videochatgpt}}                                                                               & \mr{2.87}{\textbf{BERTScore $\uparrow$}~\cite{bertscore}} \\ \cmidrule(lr){3-5} \cmidrule(lr){6-9}
                                            &                                         & 0.3                             & 0.5                                   & 0.7                               & CI                              & DO                            & CU                            & TU                           &                                          \\ \hlineB{2.5}
    \multirow{2}{*}{GPT-4o~\cite{GPT-4o}}	& Step	                                  & 15.53                           & 12.65                                 & 8.59  		                    & 3.33                            & 2.688                         & 3.559                         & 2.852                        & 0.857	                                \\
                                            & \cellcolor[gray]{0.9}Substep            & \cellcolor[gray]{0.9}19.51      & \cellcolor[gray]{0.9}16.15            & \cellcolor[gray]{0.9}11.19 		& \cellcolor[gray]{0.9}3.027      & \cellcolor[gray]{0.9}2.595    & \cellcolor[gray]{0.9}3.336    & \cellcolor[gray]{0.9}2.706   & \cellcolor[gray]{0.9}0.866	            \\ \cdashline{2-11}
    \multirow{2}{*}{InternVL2-40B-AWQ~\cite{internvl2}}          & Step	              & 15.23	                        & 12.67		                            & 8.68                              & 3.192                           & 2.606                         & 3.381	                      & 2.670		                 & 0.858	                                \\ 
                                            & \cellcolor[gray]{0.9}Substep            & \cellcolor[gray]{0.9}19.79	    & \cellcolor[gray]{0.9}16.11		    & \cellcolor[gray]{0.9}10.89        & \cellcolor[gray]{0.9}2.869	  & \cellcolor[gray]{0.9}2.559	  & \cellcolor[gray]{0.9}3.164	  & \cellcolor[gray]{0.9}2.543   & \cellcolor[gray]{0.9}0.873	            \\ \cdashline{2-11}                                            
    \multirow{2}{*}{InternVL2-8B~\cite{internvl2}}           & Step	                  & 10.01	                        & 8.21 	                                & 5.49				                & 2.791                           & 2.374                         & 3.053	                      & 2.222		                 & 0.843	                                \\  	
                                            & \cellcolor[gray]{0.9}Substep            & \cellcolor[gray]{0.9}13.09	    & \cellcolor[gray]{0.9}10.65	        & \cellcolor[gray]{0.9}7.57			& \cellcolor[gray]{0.9}2.245      & \cellcolor[gray]{0.9}2.248    & \cellcolor[gray]{0.9}2.644	  & \cellcolor[gray]{0.9}1.978   & \cellcolor[gray]{0.9}0.862	            \\ \hlineB{2.5}
    \end{tabular}}
    \vspace{-3mm}
    \caption{Experimental results on the EgoGS dataset using different VLMs. Here, CI, DO, CU, TU in GPT-Score refer to ``Correctness of Information'', ``Detail Orientation'', ``Contextual Understanding'', ``Temporal Understanding'' respectively.}
    \vspace{-3mm}
    \label{tab:different-vlm}
\end{table*}

\section{Generating descriptions for incomplete \newline action instances}
\label{sec:incomplete action}

In Section \ref{sec:experiment}, we reported results where the streaming module invoked the VLM to predict substeps and steps at the points when each action instance ended, and to predict the overall goal when the video finished. However, in real world scenarios, it is crucial to generate predictions even when an action instance or video is still in progress.

Thus, in this section, we present the prediction results for substep instances, step instances, and the entire video at various completion stages: 25\%, 50\%, 75\%, and 100\%. This analysis aims to demonstrate the OpenHOUSE's ability to provide reliable predictions for substeps, steps, and goals even before an action instance or the entire video reaches completion.

Table \ref{tab:supp_B} presents the VLM inference results on both ground truth (GT) temporal annotations and those generated by the OpenHOUSE streaming module at different completion stages. While there is a clear trend of improved performance with increased observation, the results at 50\% completion show only slight degradation compared to full observation. Note that we only have 134 test samples for goal accuracy, so a single correct or incorrect prediction can lead to approximately a 1\% fluctuation in performance, explaining the observed perturbations.

These findings highlight that OpenHOUSE can provide reliable inferences even when only partial information is available, emphasizing the potential of OpenHOUSE in real-world, online scenarios where actions are often incomplete.

\section{Dataset Analysis}
\label{sec:dataset}

\begin{figure*}
    \centering
    \includegraphics[width=0.85\linewidth, height=0.7\linewidth]{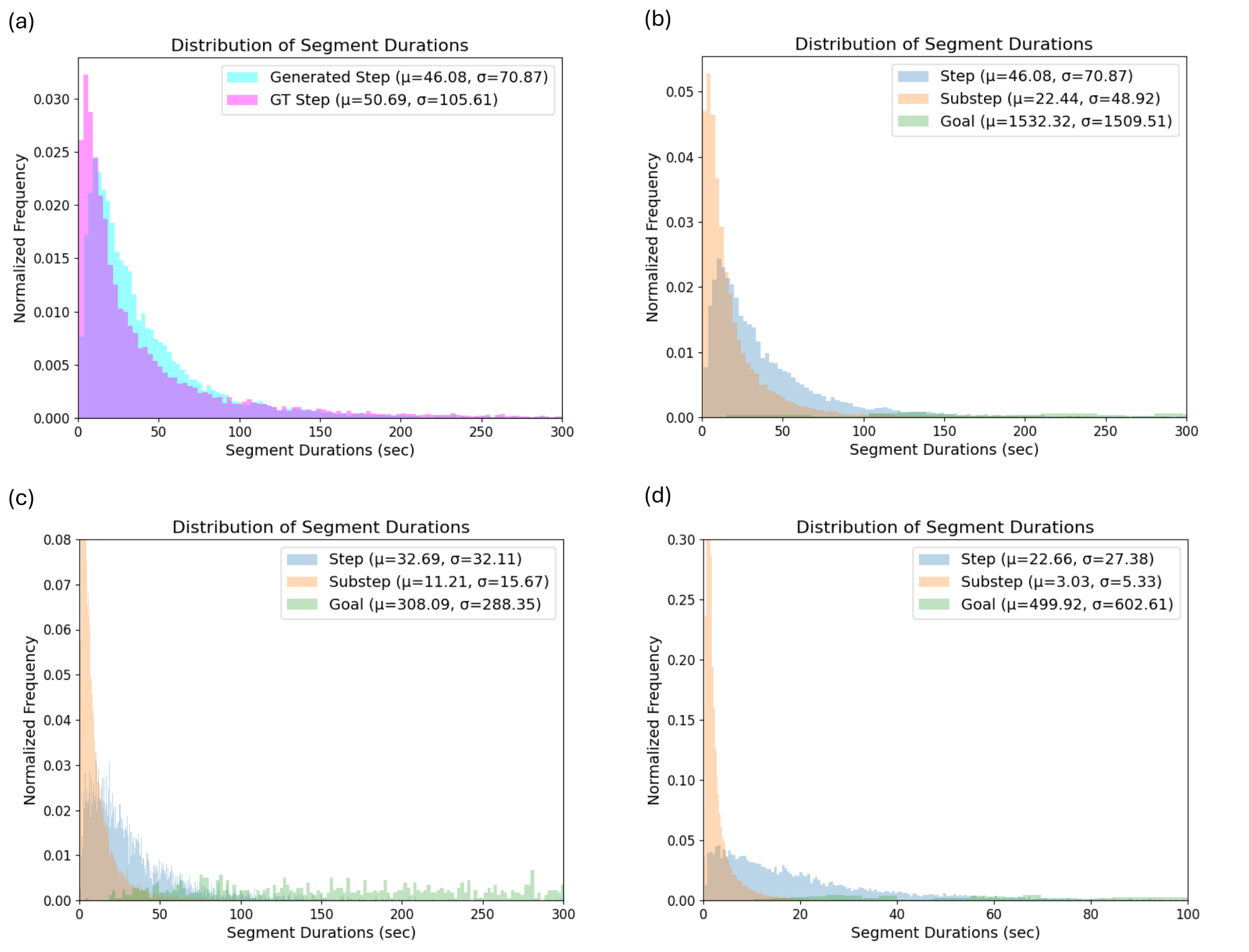}
    \caption{Dataset statistics}   
    \label{fig:egogs_pseudo}
\end{figure*}

In Section \ref{sec:dataset}, we aim to discuss the comparison between the Ego4D GoalStep (EgoGS) dataset and the Ego4D GoalStep pseudo (EgoGS pseudo) dataset generated through our dataset generation pipeline. Additionally, we provide statistics on the reconstructed hierarchical datasets, including Ego-Exo4D Keystep (EgEx) and Epic-Kitchens 100 (EK100).

\noindent\textbf{Comparison between EgoGS and EgoGS pseudo}
Figure \ref{fig:egogs_pseudo} (a) illustrates a comparison of step segment durations between EgoGS and EgoGS pseudo. The step segment duration for EgoGS averages 50.69 seconds, slightly differing from the original value reported in the Ego4D GoalStep dataset~\cite{ego4dgoalstep} because only the train and validation data, excluding the test dataset, were used for this calculation. EgoGS pseudo has an average step segment duration of 46.08 seconds, which is similar to EgoGS. Additionally, the overall distribution of step segment durations for EgoGS pseudo aligns well with that of EgoGS. This demonstrates that the dataset generated through our pipeline effectively approximates the original dataset.

\noindent\textbf{Dataset Statistics}
Figure \ref{fig:egogs_pseudo} (b), (c), (d) respectively illustrate the distributions of goal, step, and substep segment durations for EgoGS pseudo, EgEx, and EK100. The goal, step, and substep segments of EgoGS pseudo have average durations of 1532.32 seconds, 46.08 seconds, and 22.44 seconds, respectively. For EgEx, the goal, step, and substep segments have average durations of 308.09 seconds, 32.69 seconds, and 11.21 seconds, respectively. The goal, step, and substep segments of EK100 have average durations of 499.92 seconds, 22.66 seconds, and 3.03 seconds, respectively.

\noindent\textbf{Human Validators in Dataset Generation}
Five human annotators participated in the dataset annotation and validation process. Each annotator contributed approximately 20 hours of work and was compensated \$200.
\section{VLM inference details}
\label{sec:inference}

\renewcommand{\thepromptcounter}{\arabic{promptcounter}}

\renewenvironment{prompt}[1][htb]
  {\floatname{algorithm}{Prompt}
   \refstepcounter{promptcounter}
   \renewcommand{\thealgorithm}{\thepromptcounter}
   \begin{algorithm}[#1]}
  {\end{algorithm}}

\begin{prompt}
\caption{\texttt{Goal Prompt} for VLM inference.}
\definecolor{codeblue}{rgb}{0.25,0.5,0.5}
\lstset{
  backgroundcolor=\color{white},
  basicstyle=\fontsize{6.5pt}{6.5pt}\ttfamily\selectfont,
  columns=fullflexible,
  breaklines=true,
  captionpos=b,
  commentstyle=\fontsize{6.2pt}{6.2pt}\color{codeblue},
  keywordstyle=\fontsize{6.2pt}{6.2pt},
  texcl=<true|false>
}
\begin{lstlisting}[language=python]
I am planning to add annotations to a video. The annotations form a three-level hierarchy: goal, steps, and substeps. Here are the specific requirements:
1. Goal Annotation: There is only one goal annotation for the entire video.
2. Step Annotations: Each step annotation is ordered chronologically and must follow the completion of all substeps within the previous step. For example, Step 2 cannot begin until all substeps of Step 1 are completed.
3. Substep Annotations: These are specific parts of the video that detail the actions within each step.

Given an image sequence extracted from a video, predict the most appropriate goal for the video based on the frames from each step and the short form responses for each step. 
The short form responses of the steps are provided as a time-ordered list in text format, where the 0th index is the earliest step and higher indices represent more recent steps. 
Utilize this context to predict the overall goal of the video.

Generate a response:
The response should consist of a single sentence that succinctly describes the goal.
Use the following list of short form responses for each step (in text format and time-ordered):
Short form response of step: {short_form_step}

Output format must be:
Answer: (goal)
\end{lstlisting}
\label{prompt:goal}
\end{prompt}

\begin{prompt}
\caption{\texttt{Step Prompt} for VLM inference.}
\definecolor{codeblue}{rgb}{0.25,0.5,0.5}
\lstset{
  backgroundcolor=\color{white},
  basicstyle=\fontsize{6.5pt}{6.5pt}\ttfamily\selectfont,
  columns=fullflexible,
  breaklines=true,
  captionpos=b,
  commentstyle=\fontsize{6.2pt}{6.2pt}\color{codeblue},
  keywordstyle=\fontsize{6.2pt}{6.2pt},
  texcl=<true|false>
}
\begin{lstlisting}[language=python]
I am planning to add annotations to a video. The annotations form a three-level hierarchy: goal, steps, and substeps. Here are the specific requirements:
1. Goal Annotation: There is only one goal annotation for the entire video.
2. Step Annotations: Each step annotation is ordered chronologically and must follow the completion of all substeps within the previous step. For example, Step 2 cannot begin until all substeps of Step 1 are completed.
3. Substep Annotations: These are specific parts of the video that detail the actions within each step.

Given an image sequence extracted from a video clip, predict the most appropriate step occurring in the clip based on the sequence and the previous steps. 
The previous steps are provided as a time-ordered list of long form responses in text format, where the 0th index is the earliest step and higher indices represent more recent steps.
If there are no previous steps, the list will be empty. If there are more than 10 previous steps, only the 10 most recent responses will be provided. Utilize this context to improve the prediction for the current step.

First, Generate two types of responses:
Short form response: A single sentence that succinctly describes the step.
Long form response: A detailed and accurate description of the step based on the image sequence, considering the previous steps if provided.

After generating the responses, revise the long form response to ensure it aligns with the short form response for consistency.

Use the following list of previous long form responses in text format to ensure continuity and logical progression (the list may be empty if there are no prior steps, and a maximum of the 10 most recent responses will be provided):
Previous long form response: {prediction_list}

Output format must be:
Answer:
short form response: (response)
long form response (before revision): (response)
long form response (after revision): (response)
\end{lstlisting}
\vspace{3mm}
\label{prompt:step}
\end{prompt}

\begin{prompt}
\caption{\texttt{Substep Prompt} for VLM inference.}
\definecolor{codeblue}{rgb}{0.25,0.5,0.5}
\lstset{
  backgroundcolor=\color{white},
  basicstyle=\fontsize{6.5pt}{6.5pt}\ttfamily\selectfont,
  columns=fullflexible,
  breaklines=true,
  captionpos=b,
  commentstyle=\fontsize{6.2pt}{6.2pt}\color{codeblue},
  keywordstyle=\fontsize{6.2pt}{6.2pt},
  texcl=<true|false>
}
\begin{lstlisting}[language=python]
I am planning to add annotations to a video. The annotations form a three-level hierarchy: goal, steps, and substeps. Here are the specific requirements:
1. Goal Annotation: There is only one goal annotation for the entire video.
2. Step Annotations: Each step annotation is ordered chronologically and must follow the completion of all substeps within the previous step. For example, Step 2 cannot begin until all substeps of Step 1 are completed.
3. Substep Annotations: These are specific parts of the video that detail the actions within each step.

Given an image sequence extracted from a video clip, predict the most appropriate substep occurring in the clip based on the sequence and the previous substeps of the current step. 
The previous substeps are provided as a time-ordered list of long form responses in text format, where the 0th index is the earliest substep and higher indices represent more recent substeps.
If there are no previous substeps, the list will be empty. Utilize this context to improve the prediction for the current substep.

First, Generate two types of responses:
Short form response: A single sentence that succinctly describes the substep.
Long form response: A detailed and accurate description of the substep based on the image sequence, considering the previous substeps if provided.

After generating the responses, revise the long form response to ensure it aligns with the short form response for consistency.

Use the following list of previous long form responses in text format to ensure continuity and logical progression (the list may be empty if there are no prior substeps):
Previous long form response: {prediction_list}

Output format must be:
Answer:
short form response: (response)
long form response (before revision): (response)
long form response (after revision): (response)
\end{lstlisting}
\label{prompt:substep}
\end{prompt}

\noindent In section \ref{sec:inference}, we describe the details of the inference process for each level in the hierarchy: \textit{substep}, \textit{step}, and \textit{goal}, using VLM.
Additionally, we provide comprehensive details regarding inference speed measurement


\noindent\textbf{Substep Inference}
Each \textit{substep} is inferred using the following inputs: (i) video frames sampled at 1-second intervals from the corresponding \textit{substep} instance, and (ii) text predictions from prior \textit{substeps} within the same step.
With these inputs, predictions are made using the prompt \ref{prompt:substep}.
The generated short form response serves as the prediction result, while the long form response (after revision) is used as input for predicting the next \textit{substep}.

\noindent\textbf{Step Inference}
For step-level inference, the input consists of: (i) images sampled from the \textit{substep} instances within the given \textit{step} at 3.3-second intervals, and (ii) text predictions from up to 10 previous \textit{steps}. 
With these inputs, predictions are made using the prompt \ref{prompt:step}.
Again, the short form response represents the prediction result, while the long form response (after revision) is employed to predict the subsequent \textit{step}.

\noindent\textbf{Goal Inference}
\textit{Goal} inference utilizes: (i) a single image per \textit{step}, and (ii) text predictions from all the \textit{steps}. With these inputs, predictions are made using the prompt \ref{prompt:goal}.

\noindent\textbf{Inference Speed Measurement Details}
As discussed in Section 4.2, we evaluated the inference speed of a video with 2758 * 16 frames (46 minutes at 16 fps) using the Intern VL2-40B-AWQ~\cite{internvl2} model. The measurement was conducted utilizing 4 * RTX 3090 GPUs. The measured FPS represents the model's average processing FPS.

Following the technical details in~\cite{miniroad}, the reported 24 FPS includes the \textit{entire OpenHOUSE pipeline}, encompassing: (i) online feature extraction from raw frames, (ii) Streaming Module inference, and (iii) VLM inference.
\section{Implementation Details}
\label{hyperparam}

As discussed in Section 3.3.1 of the main paper, our Streaming Module consists of three heads: a state-emitting head, a progression head for steps, and a progression head for substeps. Each head shares an RNN backbone comprising three recurrent layers with a hidden state size of 768.

The state-emitting head is trained using the standard cross-entropy loss, following the approach in~\cite{actionswitch}. The progression heads for both steps and substeps are trained using the histogram loss described in~\cite{stopregressing}, configured with 10 bins and a standard deviation ($\sigma$) of 0.15.

We trained using AdamW optimizer with a learning rate of 3e-4, a batch size of 16, and a weight decay of 0.01. The model is trained for a total of 30 epochs.
\section{Streaming Module Comparison}
\label{sm}

Since our streaming module is based on the ActionSwitch~\cite{actionswitch} design, we conducted apple-to-apple evaluations on EK-100 against various class-agnostic On-TAL baselines. (Table \ref{tab:r_t2}).
Key observations are:
(i) Streaming Module (SM) without hybrid detection performs comparably to ActionSwitch, validating our design choice,
(ii) SM with hybrid strategy significantly outperforms prior methods, setting a new SOTA in class-agnostic On-TAL.
These results further confirm the effectiveness and superiority of our hybrid action boundary detection method.

\begin{table}[h]
    \centering
    \vspace{-2mm}
    \resizebox{1\linewidth}{!}{
        \renewcommand{\arraystretch}{1}
        \begin{tabular}{cccc}
        \hline
        \textbf{Method}                         & \textbf{F1@0.5}                   & \textbf{Precision@0.5}            & \textbf{Recall@0.5} \\ \hline
        CAG-QIL~\cite{cagqil}                       & 23.117                        & 21.347                        & 25.206 \\
        SimOn~\cite{simon}                          &  4.395                        &  2.351                        & 33.481 \\
        OAD-Grouping~\cite{cagqil}                  & 21.416                        & 25.533                        & 18.442 \\
        ActionSwitch~\cite{actionswitch}            & 32.444                        & 29.858                        & 35.519 \\
        SM w/o Hybrid                               & 31.459                        & 38.689                        & 26.506 \\
        \cellcolor[gray]{0.9} \textbf{SM (OpenHOUSE)}  & \cellcolor[gray]{0.9} \textbf{48.954}  & \cellcolor[gray]{0.9} \textbf{48.172}  & \cellcolor[gray]{0.9} \textbf{49.763} \\ \hline
        \end{tabular}}
    \vspace{-3mm}
    \caption{SM(Streaming Module) comparison in EK100 (Class-agnostic metrics)}
    \label{tab:r_t2}
    \vspace{-5mm}
\end{table}

{
    \small
    \bibliographystyle{ieeenat_fullname}
    \bibliography{main}
}


\end{document}